%% file: main.tex
\documentclass[runningheads]{llncs}

\usepackage{eccv}

\usepackage{eccvabbrv}

\usepackage{graphicx}
\usepackage{booktabs}    %
\usepackage{multirow}    %
\usepackage{graphicx} 
\usepackage[table]{xcolor}
\usepackage{tabularx}
\usepackage[hidelinks, hypertexnames=false]{hyperref}
\usepackage{caption}
\usepackage{subcaption}
\usepackage[accsupp]{axessibility}  %

\input{preamble}

\usepackage{orcidlink}

\begin{document}

\title{EgoPolice: A Benchmark for Egocentric Video Understanding in High-Stakes Police Body-Worn Camera Footage} 

\titlerunning{EgoPolice: A Benchmark for Egocentric Video Understanding...}

\renewcommand{\thefootnote}{\fnsymbol{footnote}}
\author{
Max Gonzalez Saez-Diez$^\star$\inst{1}\orcidlink{0009-0001-6854-6144}
Jihoon Chung\thanks{Equal Contribution \\ Project Website: \url{https://github.com/princetonvisualai/egopolice}}\inst{1}\orcidlink{0000-0002-8268-6282}\\
Adam D. Wolsky\inst{1}\orcidlink{0000-0001-6784-2593}
Gregory Lanzalotto\inst{2}\orcidlink{0009-0006-8866-0100}
Dean Knox\inst{2}\orcidlink{0000-0002-1945-7938} \\
Jonathan Mummolo\inst{1}\orcidlink{0000-0002-5639-3718}
Brandon M. Stewart\inst{1}\orcidlink{0000-0002-7657-3089}
Olga Russakovsky\inst{1}\orcidlink{0000-0001-5272-3241}
}

\renewcommand{\thefootnote}{\arabic{footnote}}

\newcommand{\olga}[1]{{\color{magenta}Olga: #1}}

\authorrunning{Max Gonzalez Saez-Diez$^\star$, Jihoon Chung$^\star$ et al.}

\institute{Princeton University \\
\email{\{saezdiez, jc5933, awolsky, jmummolo, bms4, olgarus\}@princeton.edu}\\
\and
University of Pennsylvania\\
\email{glanza@wharton.upenn.edu}, \email{dcknox@upenn.edu}}

\maketitle
\input{0_abstract}

\input{1_intro}
\input{2_related_work}
\input{3_pipeline}
\input{5_characteristics}
\input{6_results}

\input{ethics}
\input{7_conclusion}

\newpage
\section*{Acknowledgements}
This work is partially supported by the Princeton University Data-Driven Social Science, the Princeton Language and Intelligence (PLI) Initiative, the Microsoft Corporate, External, and Legal Affairs (CELA) legal diversity gift, and Princeton School of Science and Engineering awards from the Helen Shipley Hunt fund and from the Project X fund.
We also thank all the annotators for the help curating the dataset, Felix Yu and Yu Wu for help with initiating the project, and Sanghyuk Chun for valuable feedback on the paper. 

\bibliographystyle{splncs04}
\bibliography{main}
\newpage
\appendix
\begin{center}
{\Large \bfseries\boldmath Supplementary Materials for \\
EgoPolice: A Benchmark for Egocentric Video \\
Understanding in High-Stakes Police Body-Worn \\
Camera Footage\par}
\end{center}

\input{X_suppl}

\end{document}

%% file: preamble.tex
\usepackage{xspace}
\newcommand{\datasetname}{EgoPolice\xspace}

\usepackage{booktabs}
\usepackage[export]{adjustbox}
\usepackage{multirow}
\usepackage{array}
\newcommand{\boldtight}[1]{{\fontseries{b}\selectfont #1}}

\usepackage{listings}  %

%% file: 0_abstract.tex
\begin{abstract}
We introduce \datasetname, a carefully curated dataset of real, egocentric police–civilian interactions, sourced from publicly available body-worn camera videos. We select police-civilian action labels that are critical for police behavioral research and annotate them at a second-by-second granularity. The videos feature rapid and irregular camera motion, dense human interactions, and rare high-stakes events, making the dataset a challenging benchmark for motion-robust and context-aware egocentric perception. We provide two different tasks, classification and multiple-choice question-answering, and benchmark both open-source and closed-source models. We find that even the best video models like Gemini 2.5 Pro still struggle to accurately predict high-risk actions such as ``Weapon Out''. Beyond serving as a benchmark, \datasetname provides a foundation for developing models capable of identifying events of interest in large-scale body-worn camera video repositories, enabling more efficient downstream human review.

\textcolor{red}{\textbf{Content Warning:} This paper includes real police body-worn camera footage, including potentially distressing scenes.}

\keywords{Egocentric Dataset, Video Understanding, Police Body-Worn Camera Analysis}
\end{abstract}

%% file: 1_intro.tex
\section{Introduction}
\label{sec:intro}

\begin{figure}[t]
    \centering
    \includegraphics[width=\linewidth]{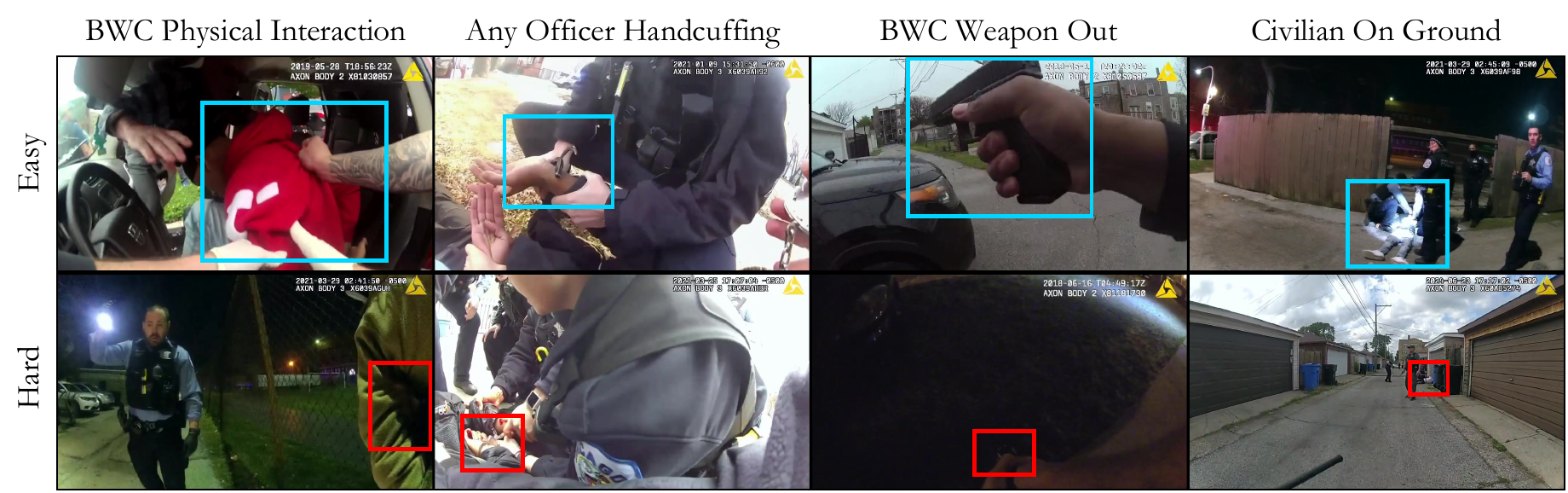}
    \caption{Sample frames of \datasetname{}. We show representative frames for \textit{Easy} (Top, \textcolor{cyan}{Cyan}) versus \textit{Hard} (Bottom, \textcolor{red}{Red}) samples. \textit{Easy} samples are correctly classified by VideoMAE V2 and CLIP~\cite{videomaev2,clip}, while \textit{Hard} samples are misclassified by both. Models perform well on clear viewpoints but struggle under low-light, occlusion, and distance.}
    \label{fig:teaser_fig1}
    \vspace{-1em}
\end{figure}

Recent advances in Vision-Language Models (VLMs) have generated significant interest in their application to a wide range of tasks and environments, including complex, high-stakes settings such as medical diagnosis, autonomous navigation, and disaster response~\cite{lu2024multimodal,zeng2025futuresightdrive,chen2026integration}. Despite their potential to support real-world decision-making, failures of these models have exposed important ethical, legal, and safety concerns \cite{grother2019face, macmillan2025arrested, attia2024effect}. These are especially pronounced in the context of law enforcement, where errors can carry civil, legal, and safety consequences. The proliferation of Body-Worn Cameras (BWCs) has generated massive archives of footage, driving demand for automated analysis. However, the data captured by these cameras differs from standard training datasets as the videos are characterized by rapid camera motion, severe occlusions, and actions that are usually filtered out in academic datasets. 
While the failure to distinguish between an innocent gesture and a real threat can carry serious social consequences, commercial vendors~\cite{axon_draft_one,truleo,abel_police,dyno_public_safety,polis_solutions} have entered the space with tools designed to interpret these records, even though the research community lacks the instruments to characterize their limitations and stress-test them against footage that is often difficult to parse even for a trained human observer~(\Cref{fig:teaser_fig1}).

To address these challenges, we introduce a carefully curated dataset of real-world, egocentric police–civilian interactions. Our work is motivated by the urgent need to benchmark and improve model reliability in first-person environments. Specifically, our primary contributions are threefold:
\begin{itemize} 
    \item \textit{Bridging the domain gap for in-the-wild video datasets:} Existing datasets focus on benign, everyday activities, yet fail to capture more nuanced and rapidly-evolving events. We construct \datasetname to provide the first benchmark for evaluating video models in egocentric law enforcement scenes.
    \item \textit{Systematically benchmarking video understanding models in high-stakes environments:} We conduct an evaluation of standard classification techniques and state-of-the-art VLMs. Our analysis reveals that despite recent advancements, models still struggle with the complexities of BWC footage.
    \item \textit{Demonstrating real-world transferability:} In ongoing work, we deploy a model trained on \datasetname in a human-in-the-loop setting to analyze a large private repository of uncurated body-worn camera videos, enabling systematic review at a scale that would be otherwise infeasible manually. This provides preliminary evidence that \datasetname can serve as a foundation for scalable police oversight tools capable of operating on real-world footage.
\end{itemize}

%% file: 2_related_work.tex
\section{Related Works}
\label{sec:formatting}
\noindent
\textbf{Police body-worn camera research.}
Police body-worn cameras (BWCs) are recording devices mounted on an officer’s chest that capture continuous video while the officer is on duty. They provide first-person documentation of police activity and were introduced with the goal of supporting police accountability and criminal investigations. Existing BWC-related research is mainly conducted by social scientists and focuses on evaluating their effectiveness, for example by measuring changes in civilian complaints, policing quality, and the civility of police–civilian encounters~\cite{JENNINGS2015480,braga2018effects,lum2020body,patterson2021there,braga2022body}. Recent advancements in machine learning have rendered the analysis of raw BWC footage computationally viable, overcoming previous constraints imposed by the sheer volume of video data. Nevertheless, current work remains focused on transcribing audible conversations and then using natural language processing techniques to study and potentially improve law enforcement behavior \cite{voigt2017, Prabhakaran2018, camp2024, camp2025, srbinovska2025}.
Apart from \cite{Hejabi2024}, which introduces a video annotation tool, and \cite{almadan2020}, which provides a face dataset based on body-worn camera footage, the analysis of the video itself remains largely unexplored, even though it offers rich context for understanding events. Nonetheless, several companies claim to analyze such footage without supporting research on their models' performance ~\cite{axon_draft_one,truleo,abel_police,dyno_public_safety,polis_solutions}. To the best of our knowledge, \datasetname is the first police BWC dataset designed to evaluate the video understanding capabilities of models,  providing a first step toward enabling independent verification and rigorous auditing of AI tools deployed in this domain.\\

\noindent
\textbf{Egocentric data/methods.}
Egocentric videos~\cite{ego4d, epickitchen2018, egoexo4d, hot3d, nymeria2024, pan2023ariadigitaltwinnew} capture videos from a first-person viewpoint, often by a head-mounted or a chest-mounted camera. Datasets used in controlled research settings typically capture actions in scripted or semi-scripted lab environments \cite{hot3d, pan2023ariadigitaltwinnew}. However, recent works have extended this to unscripted, everyday activities \cite{epickitchen2018, damen2022rescaling, ego4d}, aligning much closer with our contribution. Nonetheless, the majority of these datasets are collected in settings where camera wearers and/or people included in the footage are aware that they are being recorded for research \cite{ego4d, egoexo4d, nymeria2024}. In addition, most clips record only a few individuals and primarily capture routine activities such as cooking, playing sports, or visiting parks \cite{bansal2022viewbestviewprocedure, ego4d, epickitchen2018}.
\datasetname targets these existing gaps by proposing a dataset that uses recordings of diverse and unscripted, but also \textit{real and high-stakes} encounters between law enforcement officers and civilians. \\

\noindent
\textbf{Video annotation.} 
The development of scalable video annotation, which demands both robust infrastructure and significant human labor, has been a recurring challenge in computer vision~\cite{vondrickVidannotation, caba2015activitynet, mscoco, russakovsky2015imagenetlargescalevisual}, particularly for unscripted, real-world, in-the-wild videos that cover a huge variety of scenarios \cite{ava2018, wepdtop, epickitchen2018}. Previous work finds that minimizing cognitive load and decomposing annotation into binary subtasks are central design principles for maximizing throughput in such settings~\cite{vondrickVidannotation, sigurdsson2016}. 
To address the cost of manual labeling, recent works utilize vision models and LLMs~\cite{videopro, chen2024panda, aiannotation2026} in model-in-the-loop frameworks, combining automated predictions with human annotators to label new and more challenging data~\cite{sam2ravi,actionatlas,egoschema,motionbench}. However, LLM-assisted annotation can distort label distributions, introduce systematic errors, and propagate biases into ground truth during annotation~\cite{schroeder2025, ashwin2023usinglargelanguagemodels}. In high-stakes domains, annotator subjectivity adds additional complexity as disagreement over temporal boundaries even for well-defined classes is common~\cite{davani2021, barrett2015,Idrees2017, uma2021}. This difficulty is further compounded when footage requires multiple labels to be assigned to every frame~\cite{yeung2017}. 
We address these challenges by employing a multi-stage pipeline with objective, intent-free definitions, ensuring accurate annotations of the high-stakes dataset.\\

%% file: 3_pipeline.tex
\section{Data Collection Pipeline}
\label{sec:pipeline}

\subsection{Data Sources}\label{subsec:datasource}

\datasetname consists of police BWC footage collected from multiple sources across the United States. The dataset includes recordings released by police departments (PD), independent news networks, and a civilian oversight office. Our main source is the Chicago PD, published by the City’s Civilian Office of Police Accountability (\href{https://www.chicagocopa.org/}{COPA}). We collected all BWC videos released between July 10, 2016 and June 23, 2024. The footage includes incidents involving civilians who were critically injured or died in police custody, as well as all instances in which an officer discharged a firearm. The videos are grouped into 83 \textit{cases}. A case corresponds to an incident involving between one and 33 officers. Similarly, we collected BWC footage from five other departments (Pasadena PD, Dallas PD, Metropolitan PD of the District of Columbia, Los Angeles PD, and San Antonio PD) and two news networks, which provide recordings from additional departments. Only COPA's footage is organized into cases.

Videos from COPA and Pasadena are subject to only minor edits, involving occasional audio omissions, particularly at the beginning of videos, and the censoring of nudity. Videos from other departments are heavily edited (\eg subtitles, press-conference-style statements, slowed-down segments, or zoom-ins). We remove non-BWC segments during pre-processing, and refer the reader to the Supplementary Material for sampled frames illustrating the difference.

\datasetname does not reflect the breadth of police activity, most of which goes unrecorded or unreleased, but rather \textit{overrepresents} high-stakes police-civilian interactions, as footage containing these incidents is more likely to be released. 

\subsection{Annotation}

\begin{table*}[t]
    \centering
    \caption{Action labels and definitions in \datasetname{}.}
    \label{tab:classes}
    \renewcommand{\arraystretch}{1.3}
    \begin{adjustbox}{width=\linewidth, valign=t}
        \begin{tabular}{l l p{10cm}}
        \toprule
        \textbf{Role} & \textbf{Action} & \textbf{Definition} \\
        \midrule
        \multirow[t]{7.5}{*}{BWC Wearer}
            & \multirow[t]{2.5}{*}{Physical Interaction}
            & \textbullet~ The Body-Camera Wearer is touching the civilian. \\
            \cmidrule(lr){2-3}
            & \multirow[t]{3.5}{*}{Medical Treatment}
            & \textbullet~ BWC Wearer is touching the civilian with hands. \\[-0.5ex]
            & & \textbullet~ Both civilian and BWC Wearer's actions are visible. \\[-0.5ex]
            & & \textbullet~ BWC Wearer is directly involved in treatment. \\
            \cmidrule(lr){2-3}
            & Weapon Out
            & \textbullet~ The clip displays a firearm held by the BWC Wearer. \\
            \cmidrule(lr){2-3}
            & \multirow[t]{2}{*}{Running}
            & \textbullet~ The camera is moving/shaking substantially. \\[-0.5ex]
            & & \textbullet~ The officer is moving forwards or backwards quickly. \\
        \midrule
        \multirow[t]{5.5}{*}{Other Officer}
            & \multirow[t]{2.5}{*}{Physical Interaction}
            & \textbullet~ An officer (not the BWC wearer) is touching the civilian. \\
            \cmidrule(lr){2-3}
            & \multirow[t]{3}{*}{Medical Treatment}
            & \textbullet~ An officer (not the BWC wearer) is touching the civilian with hands. \\[-0.5ex]
            & & \textbullet~ Both civilian and police are visible. \\[-0.5ex]
            & & \textbullet~ The officer is directly involved in treatment. \\
        \midrule
        Any Police Officer
            & \multirow[t]{2}{*}{Handcuffing}
            & \textbullet~ Any officer is attempting to use or using handcuffs. \\[-0.5ex]
            & & \textbullet~ Handcuffs are clearly identifiable. \\
        \midrule
        \multirow[t]{4.5}{*}{Civilian}
            & \multirow[t]{2.5}{*}{Visibly Injured}
            & \textbullet~ The civilian is visibly injured. \\[-0.5ex]
            & & \textbullet~ The clip shows bleeding or a wound. \\
            \cmidrule(lr){2-3}
            & \multirow[t]{2}{*}{On Ground}
            & \textbullet~ The civilian is sitting, kneeling, or lying on the ground. \\[-0.5ex]
            & & \textbullet~ Ground position is clear from the video. \\
        \bottomrule
        \end{tabular}
    \end{adjustbox}
\end{table*}

\noindent
\textbf{Compiling a list of police-civilian interactions.} Our dataset captures a broad spectrum of interactions between law enforcement officers and civilians, ranging from routine conversations to high-stakes incidents such as shootings or medical emergencies. Many event types are rare due to the vast space of possible encounters. 
To address this challenge, we first compile a list of actions that are critical for policing research, explicitly distinguishing between the officer wearing the camera (BWC wearer), other officers visible in the recording who are not the recording officer (Other Officer), and civilians involved (Civilian). 

We then group these actions into classes that occur frequently enough to support computer vision research while retaining social science significance. To avoid subjective interpretation and reduce disagreement between annotators, we defined each action using objective criteria, ensuring that no prior knowledge of policing is required for annotation. \Cref{tab:classes} summarizes the action taxonomy. For handcuffing, we combine officers into ``Any Police Officer'' since multiple officers are often in close proximity, which makes it difficult to reliably distinguish individual hands. \Cref{fig:teaser_fig1} presents example frames for four different actions, illustrating both easy and challenging instances for each class.\\

\noindent
\textbf{Labeling actions.}
We designed the annotation process as a two-stage pipeline. In Stage 1, which serves as an initial filtering step, three independent annotators reviewed the candidate videos and flagged any timestamps containing visible interactions, including verbal exchanges between police officers and civilians. To ensure comprehensive coverage, we aggregated these annotations by taking the union of all identified time windows and adding a 10-second buffer before and after the start and end of each interaction. In Stage 2, the resulting interaction-rich clips were assigned to annotators for labeling across the compiled action categories at a per-second granularity. Given the subjective and high-stakes nature of the data, annotators were provided with objective action definitions that explicitly avoid inferring intent or making normative judgments. For example, the action ``BWC Wearer-Physical Interaction'' is defined as cases in which \textit{the officer wearing the camera is touching the civilian}, rather than relying on subjective interpretation. This approach ensures that action labels are based on observable behavior, reducing ambiguity for both annotators and researchers. As ``BWC Wearer-Running'' and ``BWC Wearer-Weapon Out'' can occur outside Stage 1 segments, these actions are annotated over the entire dataset. 

\begin{table*}[t]
\centering
\captionsetup{justification=centering}
\begin{minipage}[t]{0.43\textwidth}
    \centering
    \caption{Data sources.}
    \label{tab:data_sources}
    \begin{adjustbox}{width=\textwidth, valign=t}
    \begin{tabular}{l l r r}
    \toprule
    Dept. & Source & \# Vids & Time (h:mm)\\
    \midrule
    Chicago PD & COPA & 600 & 109:15 \\
    Los Angeles PD & Dept. & 82 & 2:03 \\
    D.C. PD & Dept. & 418 & 12:20 \\
    Pasadena PD & Dept. & 76 & 10:44 \\
    San Antonio PD & Dept. & 135 & 3:03 \\
    Dallas PD & Dept. & 38 & 3:00 \\
    Misc. PDs & News & 1,335 & 44:29 \\
    \midrule
    Total & & 2,684 & 184:57 \\
    \bottomrule
    \end{tabular}
    \end{adjustbox}
\end{minipage}
\hfill
\begin{minipage}[t]{0.55\textwidth}
    \centering
    \caption{Annotated actions per data source.}
    \label{tab:summary_stats_at_second_level}
    \begin{adjustbox}{width=\textwidth, valign=t}
    \begin{tabular}{lrrr||rr}
    \hline
    \multirow{2}{*}{Class} & \multicolumn{1}{c}{COPA} & \multicolumn{1}{c}{Pasadena} & \multicolumn{1}{c||}{Others} & \multicolumn{2}{c}{\datasetname{}}  \\
    & \multicolumn{1}{c}{\# sec.} & \multicolumn{1}{c}{\# sec.} & \multicolumn{1}{c||}{\# sec.} & \multicolumn{1}{c}{\# sec.} & \%  \\
    \hline
    BWC-Physical Interaction & 13,772 & 1,749 & 17,036 & 32,557 & 4.89\% \\
    BWC-Medical Treatment & 6,387 & 851 & 644 & 7,882 & 1.18\% \\
    BWC-Weapon Out & 23,961 & 1,142 & 15,063 & 40,166 & 6.03\% \\
    BWC-Running & 2,893 & 434 & 3,297 & 6,624 & 0.99\% \\
    Other-Physical Interaction & 29,167 & 2,553 & 13,223 & 44,943 & 6.75\% \\
    Other-Medical Treatment & 15,810 & 1,376 & 1,042 & 18,228 & 2.74\% \\
    Any Officer-Handcuffing & 1,138 & 152 & 2,310 & 3,600 & 0.54\% \\
    Civilian-Injured & 25,274 & 1,041 & 1,793 & 28,108 & 4.22\% \\
    Civilian-On Ground & 43,083 & 2,865 & 18,025 & 63,973 & 9.61\% \\
    \hline
    \textit{Any Class} & 75,542 & 6,122 & 49,075 & 130,739 & 19.63\% \\
    \hline
    \end{tabular}
    \end{adjustbox}
\end{minipage}
\end{table*}

We obtain second-by-second annotations by determining whether the corresponding action class is visible at any given second. Videos from COPA and Pasadena, which are of higher quality and serve as the primary data source for training and evaluation, are annotated independently by at least two annotators to ensure high-quality labels, while videos from the remaining departments are used as auxiliary training data and annotated by a single reviewer. \Cref{tab:data_sources} summarizes the data sources and \cref{tab:summary_stats_at_second_level} shows the prevalence of each annotated class stratified by origin. In total, \datasetname contains over 180 hours of BWC footage. We refer the reader to the Supplementary Material for additional statistics.

\subsection{Annotator Management}

Our Institutional Review Board determined our project to be non-human-subjects research. Given the nature of the dataset, we implemented additional safeguards, incorporating strategies rarely used in prior data collection, to protect annotators' mental health and ensure data quality. \\

\noindent
\textbf{Hiring procedure.}
To ensure the best quality, we opted to directly hire annotators from a local university, rather than using crowdsourcing platforms. 
We paid the annotators \$16/hour when the project began, which increased to \$18/hour for the final tasks. We hired 33 annotators who spent approximately 5,400 hours with the annotation cost totaling around \$92,000.
The job listing clearly stated that ``annotators that work on this project will be exposed to distressing or graphic images and sounds [...] we ask potential applicants to consider their mental and emotional ability to work with such footage before applying.'' Next, prior to hiring annotators, we interviewed each prospective annotator, again emphasizing the potential challenges to mental health due to the nature of the body camera footage. Once the annotators were hired, we asked them to sign an acknowledgment of risk form, which provided even more detail about the types of situations that may be visible in the body camera footage. 
\\

\noindent
\textbf{Mitigating vicarious traumatization.}
Research has shown that repeated exposure to violent footage can cause vicarious traumatization \cite{spence_content_mod}. We aimed to clarify this risk as early as possible, beginning with the job listing itself and reinforced again during the interview, so that annotators could make an informed decision before ever committing to the role. Prior to beginning annotation work, we provided mental health training, outlining vicarious traumatization and secondary traumatic stress and provided tools to cope with challenging video clips including taking breaks, working during the day, and self-care practices. \\

\begin{figure*}[t]
    \centering
    \begin{subfigure}[t]{0.32\textwidth}
        \centering
        \fbox{\begin{minipage}[c][3cm][c]{0.95\linewidth}
            \centering
            \includegraphics[width=\linewidth, height=3cm, keepaspectratio]{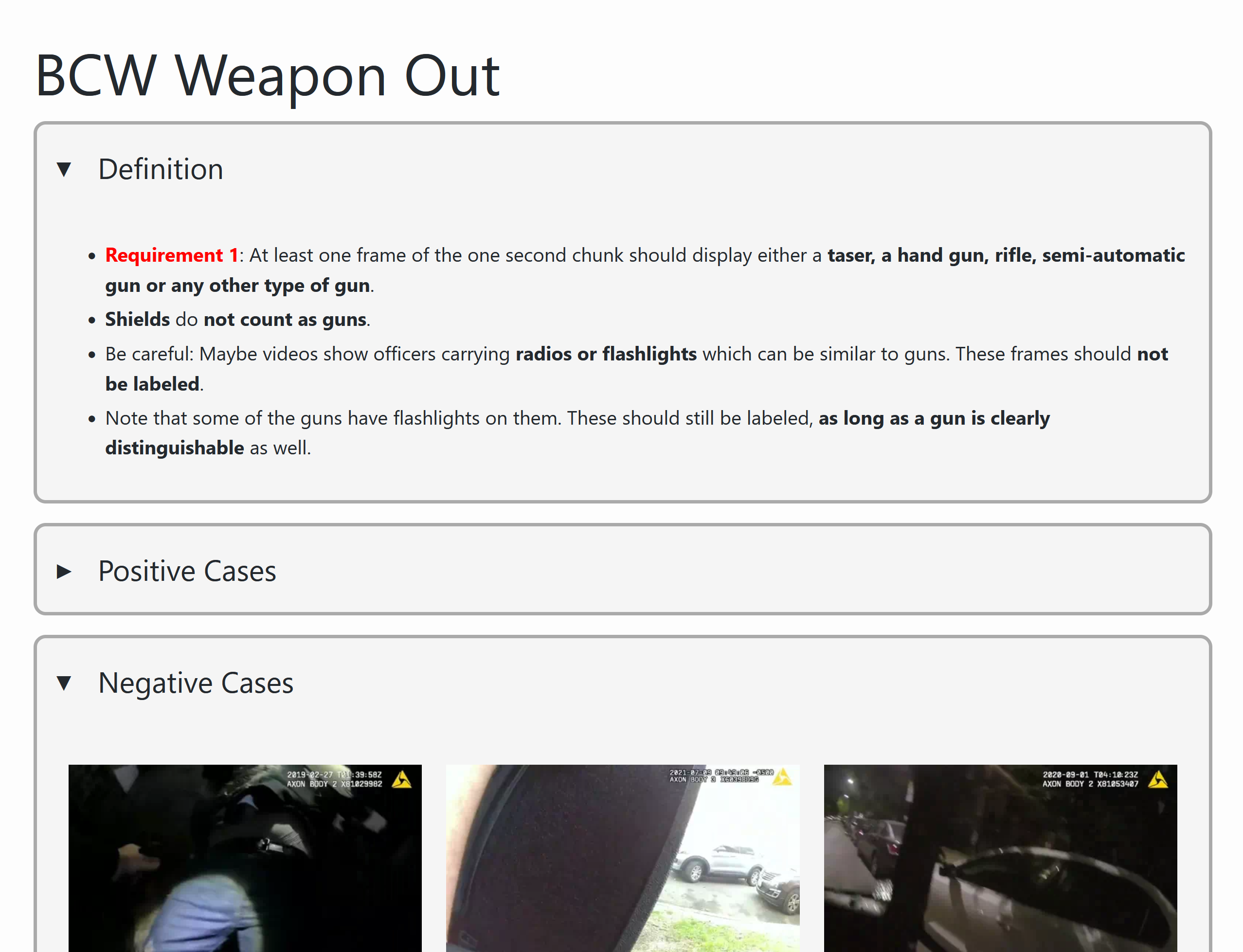}
        \end{minipage}}
        \caption{}
        \label{fig:codebook}
    \end{subfigure}
    \hfill
    \begin{subfigure}[t]{0.32\textwidth}
        \centering
        \fbox{\begin{minipage}[c][3cm][c]{0.95\linewidth}
            \centering
            \includegraphics[width=\linewidth, height=3cm, keepaspectratio]{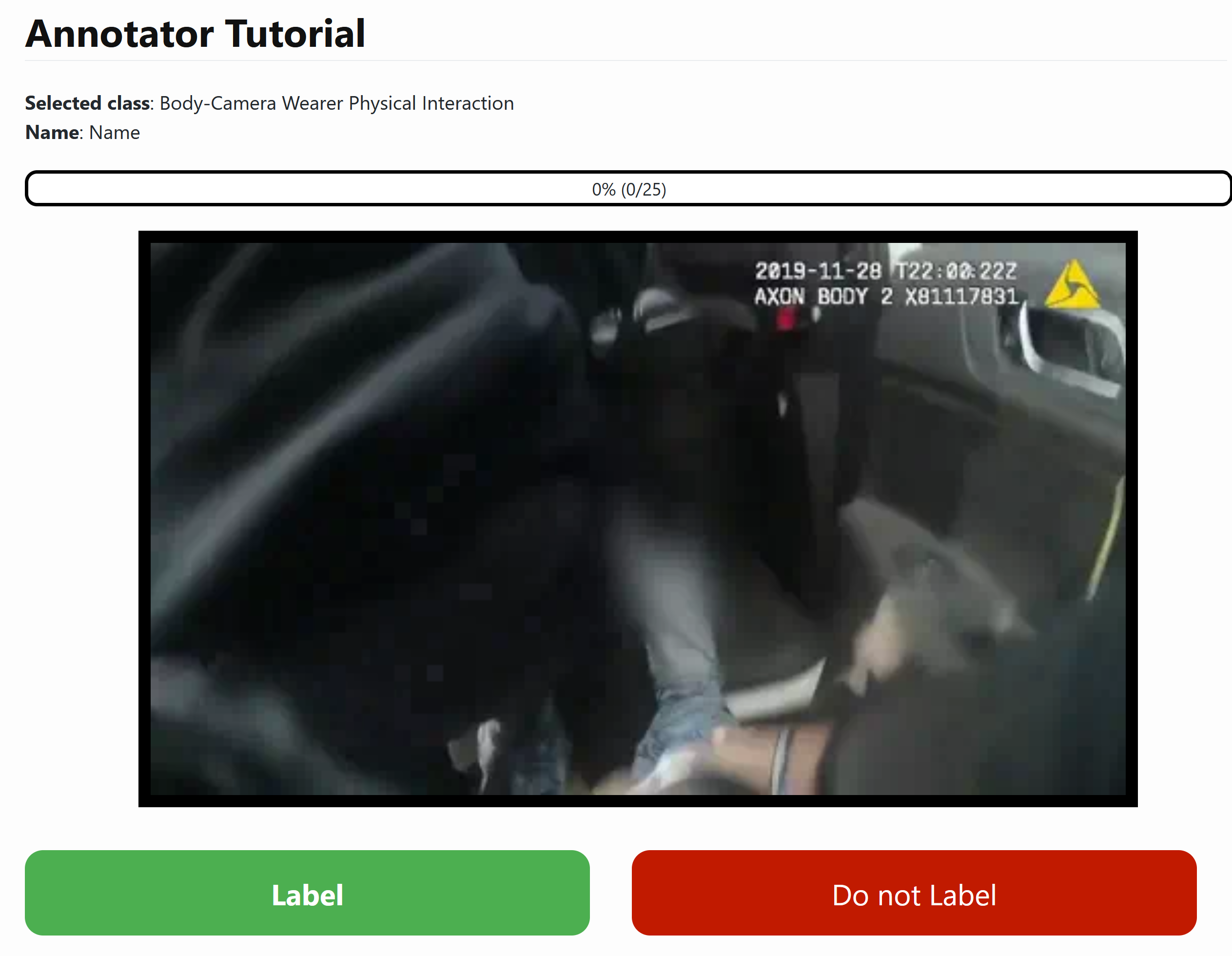}
        \end{minipage}}
        \caption{}
        \label{fig:training}
    \end{subfigure}
    \hfill
    \begin{subfigure}[t]{0.32\textwidth}
        \centering
        \fbox{\begin{minipage}[c][3cm][c]{0.95\linewidth}
            \centering
            \includegraphics[width=\linewidth, height=3cm, keepaspectratio]{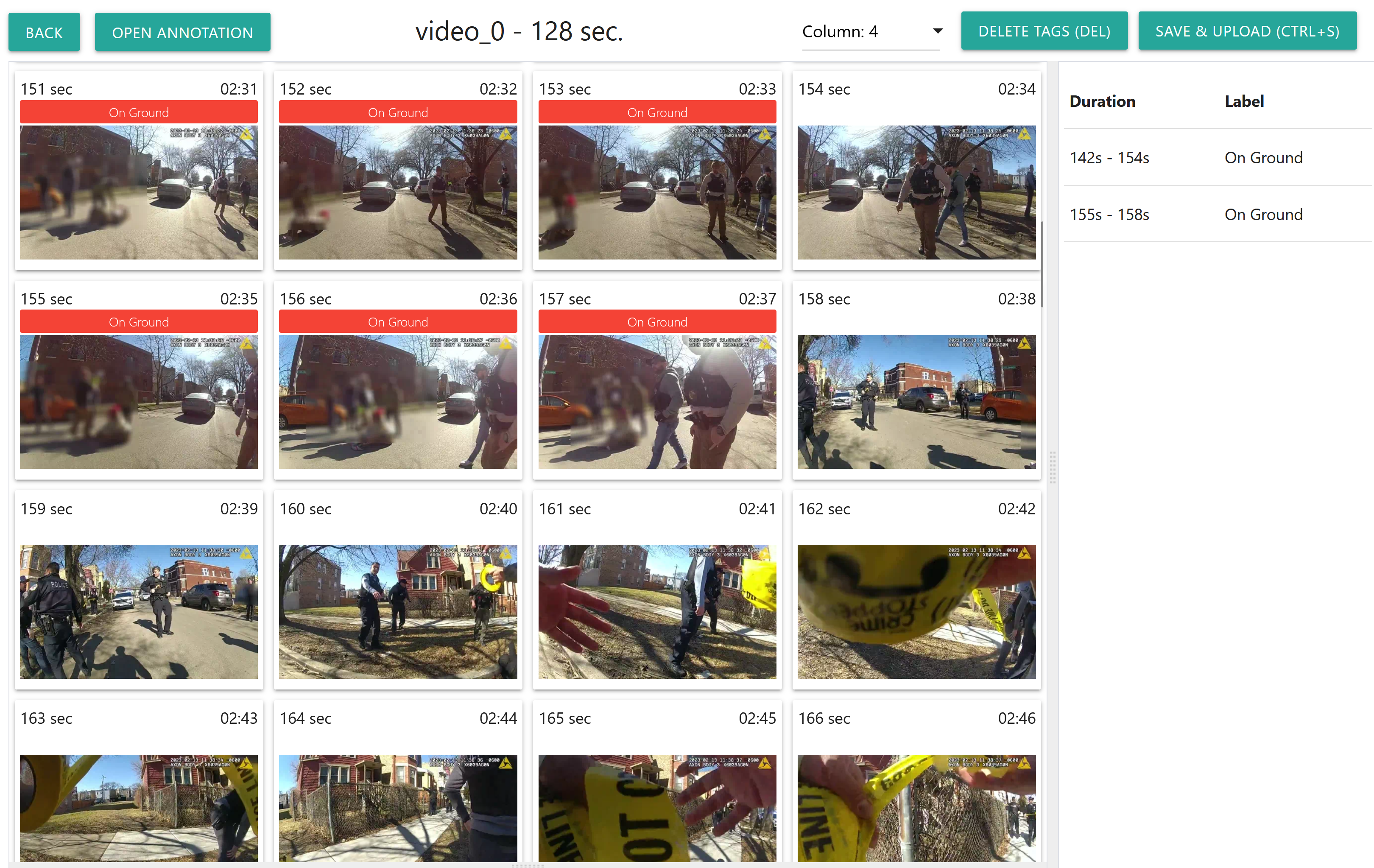}
        \end{minipage}}
        \caption{}
        \label{fig:bwcannotation}
    \end{subfigure}
    \caption{Annotation Interface. (a) Excerpt from the document providing action definitions with paired positive and negative examples. (b) Training module with practice clips. (c) Custom web-based annotation tool to support efficient labeling.}
    \label{fig:tools}
\end{figure*}

\noindent
\textbf{Initial training.}
We provided structured onboarding and conducted practice annotations with feedback to ensure consistency across labels. 
We compiled a document (\cref{fig:codebook}) that included definitions along with positive and negative examples for every class. After reviewing the definitions, annotators completed a tutorial (\cref{fig:training}) consisting of 25 practice annotations. The tutorial served as an assessment with detailed feedback explaining inclusion and exclusion criteria. 
\\

\noindent
\textbf{Annotation tool.}
We reviewed available off-the-shelf annotation tools, but ultimately decided to develop our own to ensure quality and speed of annotation. Our tool allows both managers to assign video clips to annotators, and annotators to have password-protected access to their unique assignments. Annotators can easily click-and-drag across multiple one-second frames of a video, which allows for a faster annotation process than existing tools that require multiple clicks to annotate a single time window. Unlike existing video annotation tools where annotators often have a single large video player in the center of the screen, we found that showing a grid of 1-second clips that each loop led to much faster annotation without increase in annotation errors. Additionally, we found less discomfort from observing disturbing scenes when the video is presented as a small grid versus being played in a full-sized video player. Although the tool supports multi-label annotation,  we assigned the annotators one action item at a time when labeling. Throughout our process, we found that annotators often labeled incorrectly or omitted labels due to complexity and annotation fatigue when given multiple actions to label simultaneously. \Cref{fig:bwcannotation} shows a screenshot of our annotation tool which is developed with PHP and JavaScript, using MySQL as the main database. The tool's source code is available on the project website, allowing anyone to download and host it on their own web server.
\\

\noindent
\textbf{Feedback and monitoring.}
We implemented ongoing mental health check-ins and close monitoring to support annotators' long-term well-being. In parallel, we continuously monitored annotation quality by reviewing annotations from each annotator to ensure adherence to the action definitions. When we found deviations, we provided targeted feedback, retraining, and retroactive correction of affected annotations. Furthermore, we manually verified approximately 25\% of all annotations and viewed every video in the dataset. We find that our mean inter-annotator agreement (Krippendorff's $\alpha$~\cite{krippendorff}) is 79.4\%.

%% file: 5_characteristics.tex
\section{Data Challenges}
\label{sec:data_challenges}

The characteristics of BWC footage can make action recognition more challenging than in conventional video benchmarks. We highlight two properties of \datasetname{} that contribute to this difficulty: severe camera motion during annotated actions and limited discriminative signal from global appearance. \\

\noindent
\textbf{Egocentric Camera Motion.} Most egocentric datasets capture routine, low-stakes activities~\cite{ego4d,epickitchen2018}. These datasets are designed for machine learning research, so actions are typically recorded clearly, with minimal camera movement and the camera generally pointed at the target. In real-world settings, however, actions can occur amid significant motion from either the subject or the camera, and in high-stakes environments such as policing, events like foot pursuits or physical altercations are inherently associated with extreme camera motion.

\begin{figure}[!t] 
    \centering
     \includegraphics[width=\columnwidth]{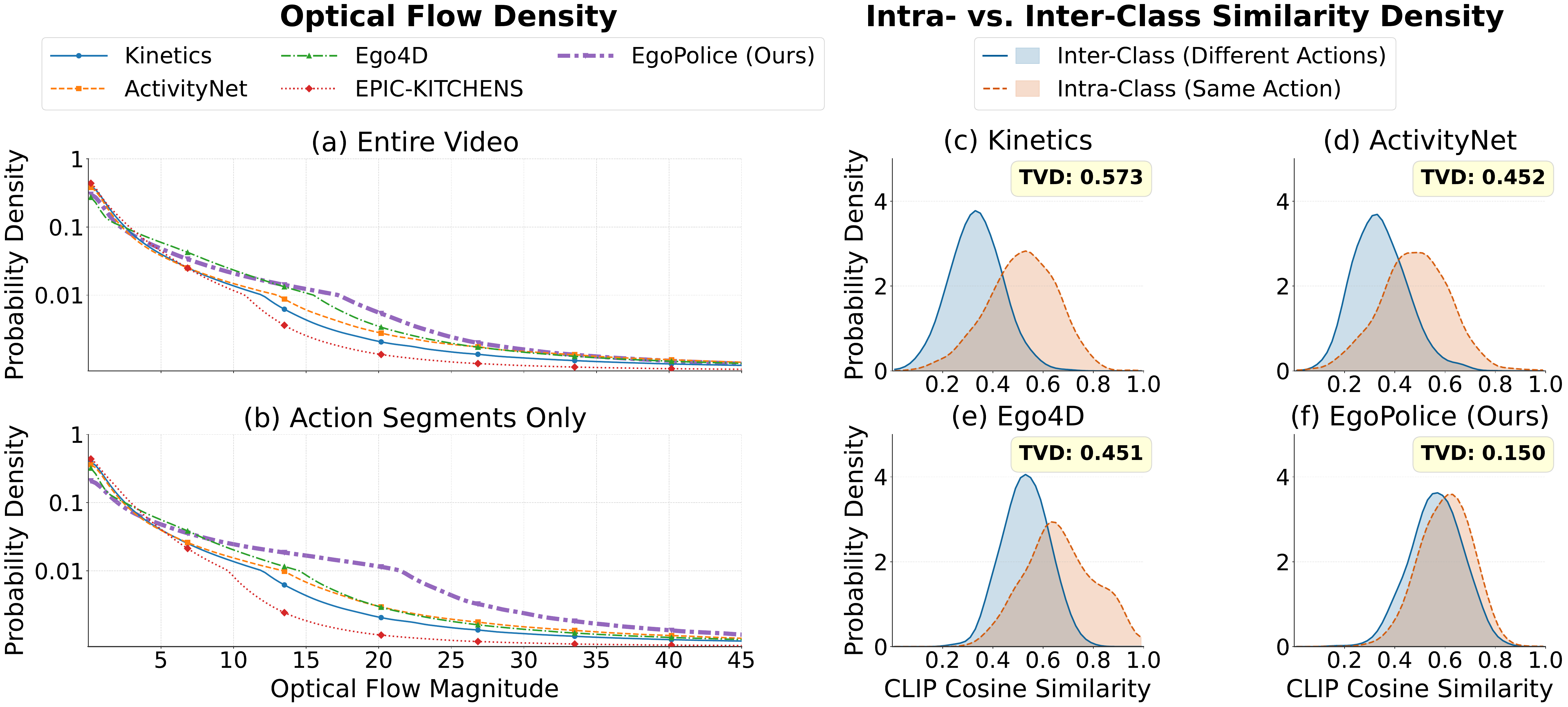} 
    \caption{Comparisons with existing datasets. \textbf{(a-b)} \datasetname{} shows higher motion overall, which is more noticeable around segments where actions are labeled. \textbf{(c-f)} Distribution of CLIP cosine similarity between frames from the same class and frames from different classes. \datasetname{} shows similar distributions for both, while other datasets show higher similarity when frames are sampled from the same class.}
    \label{fig:optical_flow}
\end{figure}

To quantitatively compare the amount of motion across datasets, we compute optical flow magnitude using FastFlowNet-v2~\cite{Kong_2021_ICRA}. \Cref{fig:optical_flow} (a,b) shows a density plot of optical flow magnitudes for \datasetname and other video datasets~\cite{kay2017kineticshumanactionvideo,caba2015activitynet,ego4d,epickitchen2018}.
When computed over the entire dataset (\cref{fig:optical_flow}a), \datasetname exhibits only slightly more motion than Ego4D~\cite{ego4d}, another standard in-the-wild egocentric dataset. Differences emerge when motion statistics are computed specifically on annotated action segments (\cref{fig:optical_flow}b). In existing datasets, optical flow magnitudes drop off quickly during action segments, reflecting that the camera is typically stationary while interacting with objects or completing tasks. In contrast, \datasetname shows a heavy-tailed distribution, with substantially more probability mass at high flow values ($>20$). This highlights a central challenge in our domain: the moments that are most semantically important contain severe visual degradation, including blur, large displacements, and rapid viewpoint shifts.\\

\noindent
\textbf{Visual Cue Diversity.} In many standard video benchmarks, action classes are strongly tied to backgrounds or overall scenes (\eg, ``playing soccer'' and ``playing cards''). As a result, models can often rely on the global scene context to classify an action. In contrast, the high-stakes scenarios in \datasetname{} unfold in far more visually uniform settings such as streets, hallways, and vehicle interiors.

We illustrate this challenge in \cref{fig:optical_flow} (c-f) by plotting the cosine-similarity distributions of CLIP embeddings~\cite{clip} for pairs of random frames sampled either within the same action class or across different classes. In datasets like Kinetics~\cite{kay2017kineticshumanactionvideo} and ActivityNet~\cite{caba2015activitynet}, these distributions differ substantially, reflected in high Total Variation Distance (TVD) scores ($0.573$ and $0.452$). This indicates that frames from the same class tend to be noticeably more similar to each other than to frames from other classes. This property allows models to rely on scene context and exploit shortcuts to classify actions. 
In contrast, for \datasetname, the two distributions almost overlap perfectly, with a TVD of only $0.150$.  
Moreover, both distributions are shifted toward higher similarity values, around 0.6, which is substantially higher than Kinetics and ActivityNet in particular. This combination of low inter-class separability and high overall visual similarity indicates that global appearance cues can offer little discriminative signal for the model to work with. 
As a result, models must rely on genuine contextual understanding and fine-grained visual cues to correctly determine the visible action.

%% file: 6_results.tex
\section{Experiments}
\label{sec:experiments}

To better understand how models transfer to the BWC domain, we evaluate \datasetname{} under two paradigms reflecting dominant approaches in video understanding. These are supervised classification, which probes representation quality under task-specific training, and multiple-choice question answering (MCQ) with video-language models, which assesses zero-shot reasoning. Together, these experiments quantify how well existing models handle this challenging setting.

\subsection{Linear probing for action classification}
\label{subsec:classification}

\noindent
\textbf{Model families and training protocols.}
Training strong video models from scratch typically requires orders of magnitude more data than we can feasibly collect for the BWC domain. Instead we evaluate how well representations transfer to this setting. To this end, we use standard embedding models \cite{hiera, dinov2, xclip, clip, videomaev2} as frozen feature extractors, training lightweight classification heads on top of the resulting representations. For models with a \texttt{CLS} token, we use the token as the global frame-level representation. For models that do not use a \texttt{CLS} token, we compute the 1-second-level representation by mean-pooling the frame-level features, sampling frames at 16 fps. Training follows the linear-probing protocol of DINOv2~\cite{dinov2}. We evaluate models under 1-second, 10-second and 1-minute window settings, where a window is labeled positive if the action happens at least once. For evaluation, predictions are made independently for each second and aggregated over each window by taking the maximum score. We report both class-level F1 and mean F1 score averaged across classes in \cref{tab:comprehensive_results}. \\

\noindent
\textbf{Data splits and generalization.} We evaluate classification performance using 6-fold cross-validation. As described in \cref{sec:pipeline}, videos from COPA and Pasadena are minimally edited and thus best reflect real-world BWC footage. Videos from the remaining departments are often heavily edited (\eg, with overlaid subtitles), making them unsuitable as clean evaluation data. For this reason, we restrict our cross-validation splits to COPA videos. All COPA \textit{cases} recorded \textit{before} 2024 are randomly split into six folds, ensuring each fold contains a similar number of police cases. For each fold, one split is used as the test set (called the \textbf{in-distribution (ID)} test set), while the remaining cases are split into training and validation sets with a 4:1 ratio. Splitting is performed at the case level to prevent leakage across splits (as cases are constituted by a varying number of videos of the same event recorded by different officers). Videos from the remaining departments are treated as auxiliary training data and are added to each fold's training set as augmentation, but are not included in validation or test splits. In addition, we evaluate generalization under two out-of-distribution settings. First, \textbf{Out-of-distribution Time (OT)} includes COPA videos recorded in 2024, measuring temporal generalization since these videos are recorded in the same location but later in time than the data in the training set. Second, \textbf{Out-of-distribution Location (OL)} includes videos from the Pasadena Police Department, measuring geographic and departmental transfer.

\begin{table*}[t]
    \begin{minipage}[t]{0.57\textwidth}
        \makeatletter\def\@captype{table}\makeatother
\caption{Linear probing classification results.}
\label{tab:comprehensive_results}
\resizebox{\linewidth}{!}{
\begin{tabular}{lrrrrrrrrr}
\toprule
 & \multicolumn{3}{c}{1sec} & \multicolumn{3}{c}{10sec} & \multicolumn{3}{c}{1min} \\
\cmidrule(lr){2-4} \cmidrule(lr){5-7} \cmidrule(lr){8-10}
Model & \multicolumn{1}{c}{ID} & \multicolumn{1}{c}{OT} & \multicolumn{1}{c}{OL} & \multicolumn{1}{c}{ID} & \multicolumn{1}{c}{OT} & \multicolumn{1}{c}{OL} & \multicolumn{1}{c}{ID} & \multicolumn{1}{c}{OT} & \multicolumn{1}{c}{OL} \\
\midrule
\textit{Random Baseline} & 7.9 & 13.9 & 6.7 & 11.3 & 20.2 & 8.4 & 18.1 & 32.3 & 13.3 \\
Hiera \cite{hiera} & 34.2 & 40.9 & 26.1 & 44.8 & 48.9 & 33.4 & 48.8 & 56.4 & 35.6 \\
DINOv2 \cite{dinov2} & \boldtight{40.2} & \underline{43.9} & 25.7 & \underline{49.4} & \underline{50.5} & 35.3 & 54.2 & \underline{56.9} & 42.0 \\
X-CLIP \cite{xclip} & 36.2 & 39.9 & \underline{30.8} & 49.3 & 48.3 & \underline{45.8} & \underline{55.4} & 54.7 & \boldtight{51.8} \\
CLIP \cite{clip} & \underline{39.2} & 42.1 & 28.5 & 48.5 & 49.8 & 40.6 & 53.5 & 54.9 & 46.1 \\
VideoMAE V2 \cite{videomaev2} & 37.9 & \boldtight{48.5} & \boldtight{30.9} & \boldtight{51.8} & \boldtight{59.3} & \boldtight{46.2} & \boldtight{58.4} & \boldtight{63.7} & \underline{49.8} \\
\bottomrule
\end{tabular}
}
\vspace{0.1pt}\\
\resizebox{\linewidth}{!}{
\begin{tabular}{lrrrrrrrrr}
\multicolumn{10}{c}{\textbf{Per-Class F1 Breakdown (ID, 1s)}} \\
\midrule
& \multicolumn{4}{c}{\textit{BWC}} & \multicolumn{2}{c}{\textit{Oth. Off.}} & \multicolumn{1}{c}{\textit{Any}} & \multicolumn{2}{c}{\textit{Civ.}} \\
\cmidrule(lr){2-5} \cmidrule(lr){6-7} \cmidrule(lr){8-8} \cmidrule(lr){9-10}
Model & \multicolumn{1}{c}{MT} & \multicolumn{1}{c}{PI} & \multicolumn{1}{c}{Run} & \multicolumn{1}{c}{Wpn} & \multicolumn{1}{c}{MT} & \multicolumn{1}{c}{PI} & \multicolumn{1}{c}{Cuf} & \multicolumn{1}{c}{Inj} & \multicolumn{1}{c}{Gnd} \\
\midrule
{Random Baseline} & 2.9 & 5.9 & 0.9 & 12.2 & 6.9 & 12.3 & 0.6 & 11.2 & 18.2 \\
CLIP \cite{clip} & \boldtight{40.0} & \underline{63.4} & \underline{2.9} & \underline{30.7} & \boldtight{36.4} & \boldtight{52.6} & \underline{4.1} & \boldtight{56.3} & \boldtight{66.0} \\
VideoMAE V2 \cite{videomaev2} & \underline{31.2} & \boldtight{63.7} & \boldtight{15.0} & \boldtight{40.0} & \underline{25.1} & \underline{49.7} & \boldtight{15.8} & \underline{39.7} & \underline{61.2} \\
\bottomrule
\end{tabular}
}
    \end{minipage}
    \hfill
    \begin{minipage}[t]{0.43\textwidth}
        \makeatletter\def\@captype{table}\makeatother
        \caption{Zero-shot VLM results.}
        \label{tab:llm_mcq}
        \resizebox{\linewidth}{!}{
            \begin{tabular}{lcrrr}
            \toprule
            Model & param & \multicolumn{1}{c}{1sec} & \multicolumn{1}{c}{10sec} & \multicolumn{1}{c}{1min} \\
            \midrule
            \; \textit{Random Baseline}                        & -  & 20.0 & 20.0 & 20.0 \\
            \textit{\textbf{Proprietary Models}} \\
\; GPT-4.1-Nano~\cite{gpt4}             & -  &            18.6  &            18.7  &            10.8  \\
\; Gemini 2.5 Flash~\cite{gemini25}     & -  &            65.8  &            69.7  &            66.1  \\
\; GPT-4.1~\cite{gpt4}                  & -  &    \textbf{77.7} & \underline{76.2} & \underline{74.4} \\
\; Gemini 2.5 Pro~\cite{gemini25}       & -  & \underline{76.9} &    \textbf{77.5} &    \textbf{76.9} \\
            \midrule
            \textit{\textbf{Open-source Models}} \\
                \; LLaVA-Mini~\cite{llavamini}          & 8B & 15.1 & 15.7 & 15.5 \\
                \; LLaMA-VID~\cite{li2024llamavid}      & 7B & 38.1 & 37.7 & 30.2 \\
                \; VideoLLaVA~\cite{lin2024video}       & 7B & 39.4 & 35.6 & 31.8 \\
                \; MERV~\cite{merv}                     & 7B & 43.0 & 41.2 & 39.5 \\
                \; CogVLM2-Video~\cite{hong2024cogvlm2} & 8B & 46.4 & 48.7 & 43.8 \\
                \; Qwen2.5-VL~\cite{qwen25}              & 7B & 52.5 & 55.8 & 55.5 \\
                \; VideoLLaMA3~\cite{videollama3}       & 7B & \underline{58.0} & \underline{61.1} & \underline{58.5} \\
                \; InternVL3~\cite{internvl3}           & 8B & \textbf{64.7} & \textbf{64.5} & \textbf{60.6} \\
            \bottomrule
            \end{tabular}
        }
    \end{minipage}
\end{table*}

\subsection{Zero-shot evaluation of Video-LLMs}
\label{subsec:mcq}

In addition to trained classifiers, we evaluate video-language models on \datasetname in a zero-shot multiple-choice question answering setting. This probes semantic understanding and temporal reasoning without task-specific fine-tuning.
We construct a benchmark of 12{,}000 questions, with 500 questions per action (including \textit{None of the above}) on 1-second and 10-second-long clips and 200 questions per action on 1-minute-long clips. Each question consists of one correct answer and four incorrect choices. Wrong action choices are uniformly randomly selected among actions that never happened in the clip, with the last option always being \textit{None of the above}.
For image-based VLMs~\cite{gpt4,gemini25}, we sample frames at 1 fps with a minimum of 8 frames and provide them as image inputs. For video-based VLMs~\cite{llavamini,li2024llamavid,lin2024video,merv,hong2024cogvlm2,qwen25,videollama3,internvl3}, we follow the author-recommended video sampling strategies. All models are evaluated in a zero-shot setting using a fixed prompt template shared across models. We only use videos from COPA and Pasadena. 

\subsection{Results}
\label{subsec:results}

\noindent
\textbf{Quantitative analysis.} \Cref{tab:comprehensive_results} and \cref{tab:llm_mcq} display the performance of the trained classification models and the zero-shot VLM models, respectively.

In the classification setting, we find VideoMAE V2~\cite{videomaev2} and X-CLIP~\cite{xclip}, both trained on video data, to be the best-performing models outperforming CLIP \cite{clip} and DINOv2 \cite{dinov2} which are trained on images. These results suggest that video pretraining is useful and allows for a more robust transfer to unseen situations.
\Cref{tab:comprehensive_results} (bottom) shows that models perform well on classes such as ``Civilian-Injured'' and ``Civilian-On Ground'', due to strong visual cues such as visible injuries.
However, performance is substantially lower for ``Any Officer-Handcuffing''. This is partly due to the rarity of this class relative to others in the dataset, and partly because the task requires detecting handcuffs that are often small and partially occluded. Additionally, we observe that CLIP beats VideoMAE V2 on some appearance-grounded classes while VideoMAE V2 leads on motion-dependent ones, suggesting that robust BWC action recognition requires combining appearance and motion-based representations.

In the zero-shot task, we see LLaVA-Mini~\cite{llavamini} and GPT-4.1-Nano~\cite{gpt4} performing worse than random guessing due to outputting captions instead of answering the corresponding letter of the correct choice or strong tendencies to choose \textit{None of the above} option. 
There is a tendency for some models to perform better on 10-second clips than on 1-second clips. We believe that 1-second clips may be too short for VLMs to correctly analyze the action happening in the video, while 10-second clips provide some temporal context that the model can use to infer the correct label. Meanwhile, with 1-minute clips, the task turns into a \textit{needle-in-the-haystack} problem, leading to a reduction in accuracy. 
Among the models we have tested, we find Gemini 2.5 Pro~\cite{gemini25} to perform the best. However, given its performance is only 76.9\% on 1-minute video clips, we argue that this model is not yet suitable for autonomous, real-world deployment. \\

\begin{figure}[t]
    \centering
    \captionsetup[subfigure]{belowskip=2pt}
    \begin{subfigure}{0.49\textwidth}
        \centering
        \includegraphics[width=\textwidth]{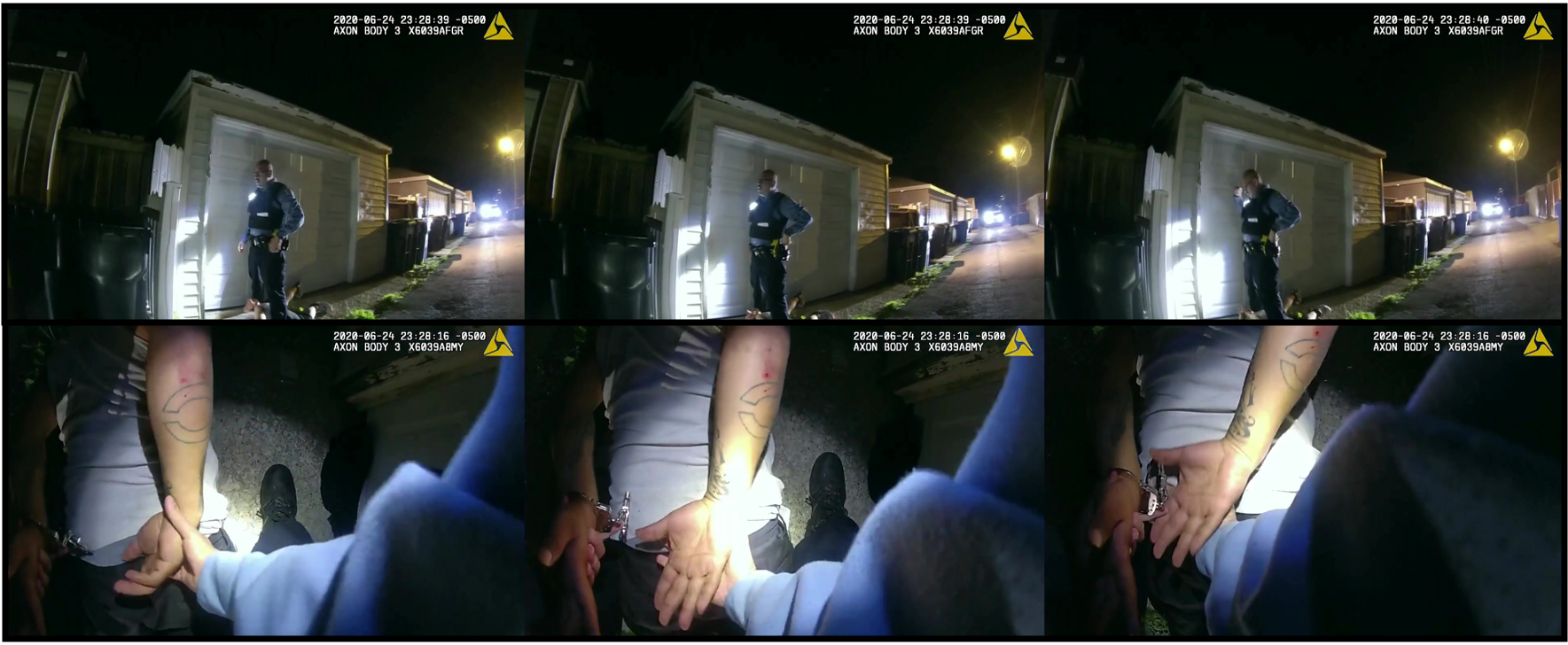}
        \caption{Low-light conditions}
        \label{minifig:lowlight}
    \end{subfigure}
    \hfill
    \begin{subfigure}{0.49\textwidth}
        \centering
        \includegraphics[width=\textwidth]{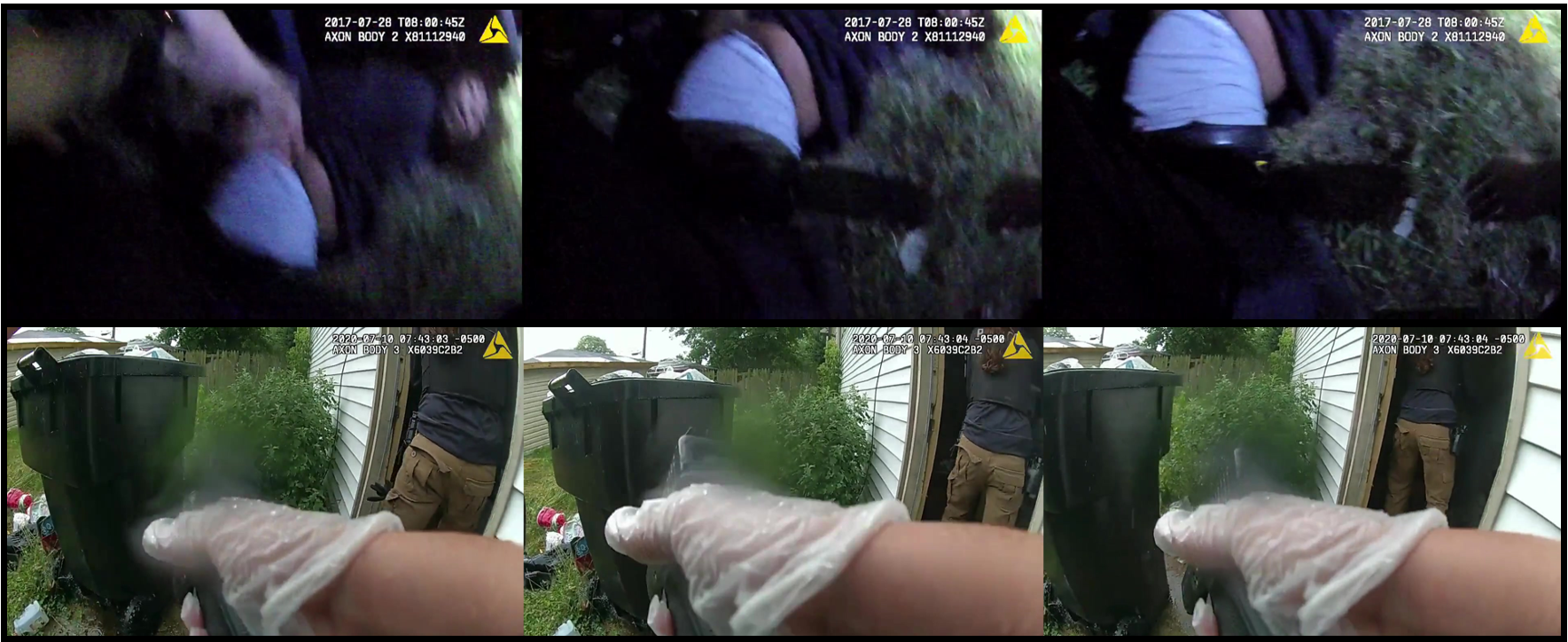}
        \caption{Blur}
        \label{minifig:blur}
    \end{subfigure}

    \begin{subfigure}{0.49\textwidth}
        \centering
        \includegraphics[width=\textwidth]{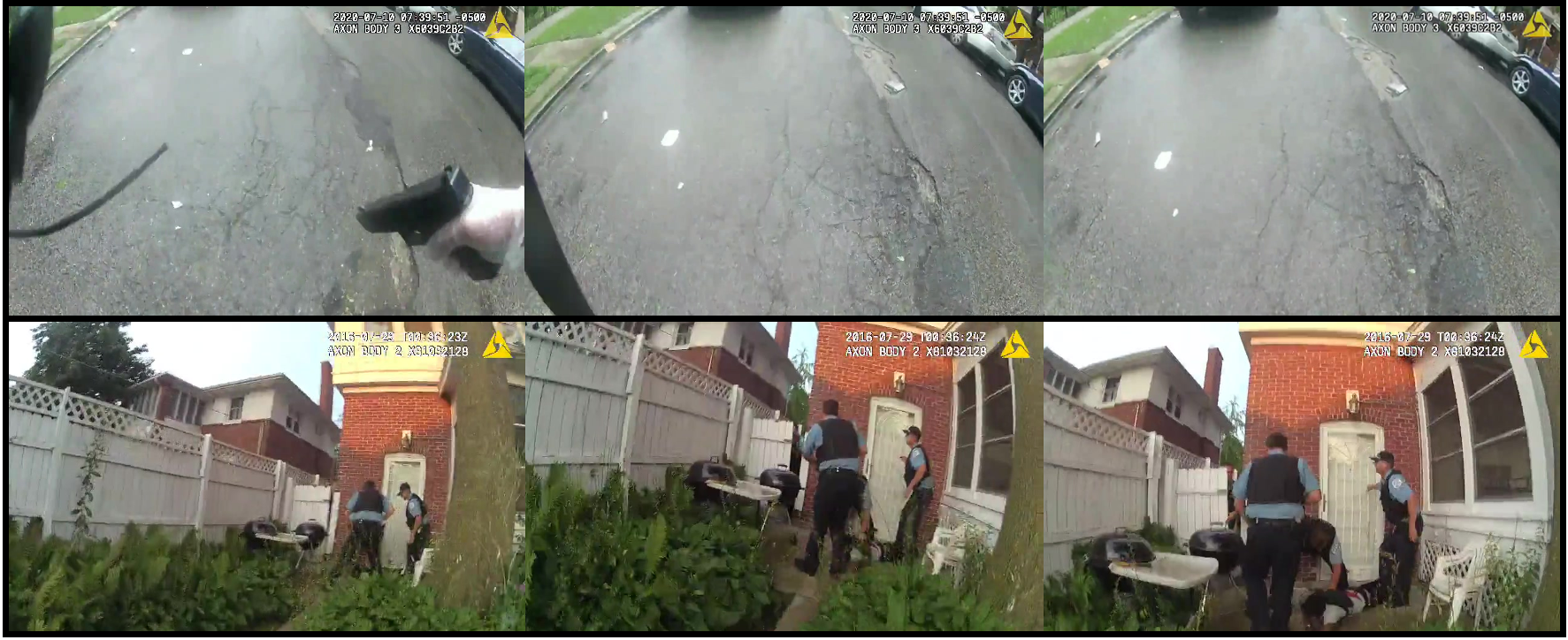}
        \caption{Fast occurrence}
        \label{minifig:fasto}
    \end{subfigure}
    \hfill
    \begin{subfigure}{0.49\textwidth}
        \centering
        \includegraphics[width=\textwidth]{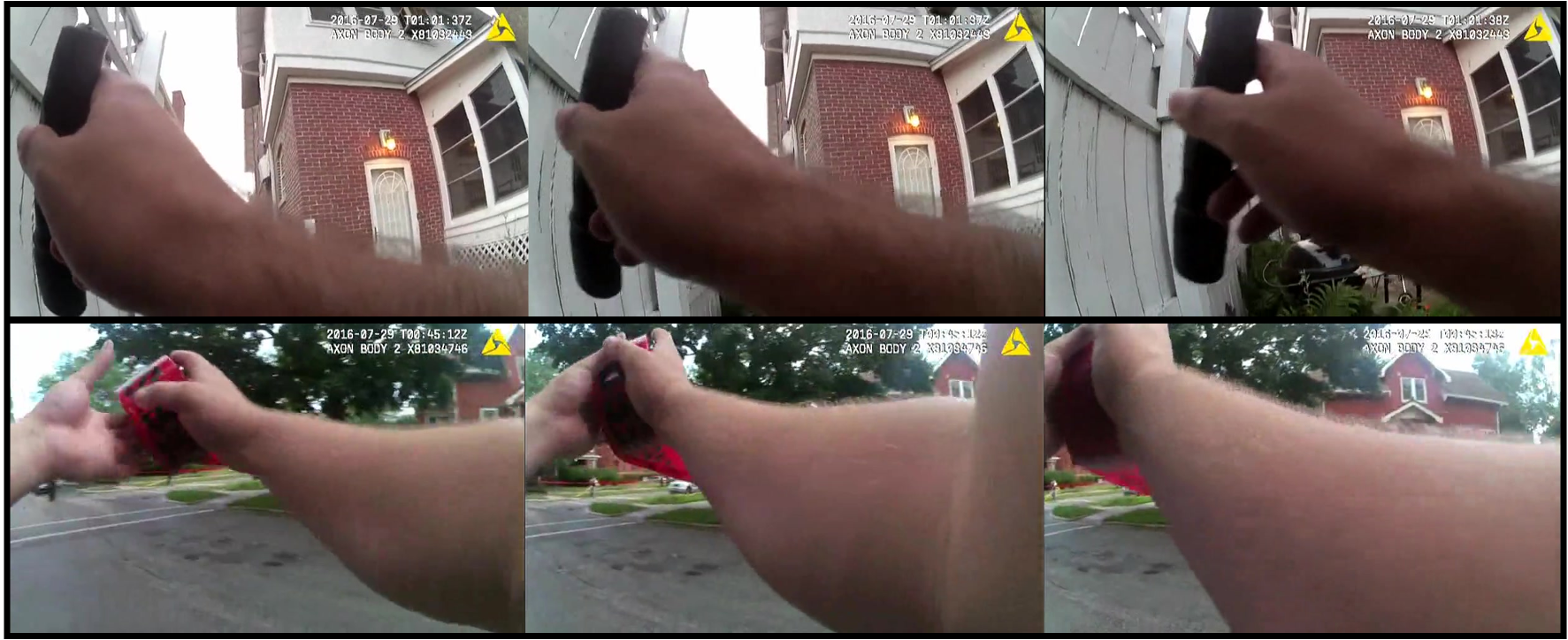}
        \caption{Misidentification}
        \label{minifig:misid}
    \end{subfigure}

    \begin{subfigure}{0.49\textwidth}
        \centering
        \includegraphics[width=\textwidth]{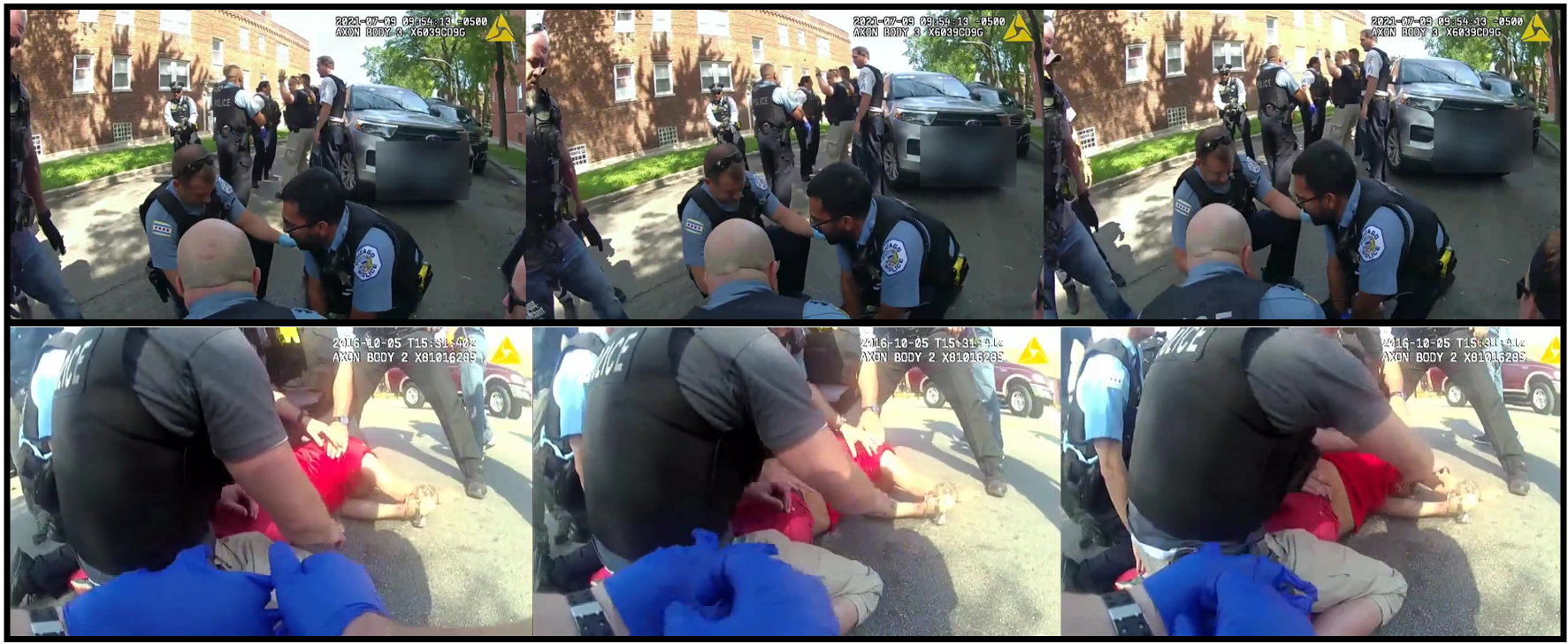}
        \caption{Spurious correlation}
        \label{minifig:scorr}
    \end{subfigure}
    \hfill
    \begin{subfigure}{0.49\textwidth}
        \centering
        \includegraphics[width=\textwidth]{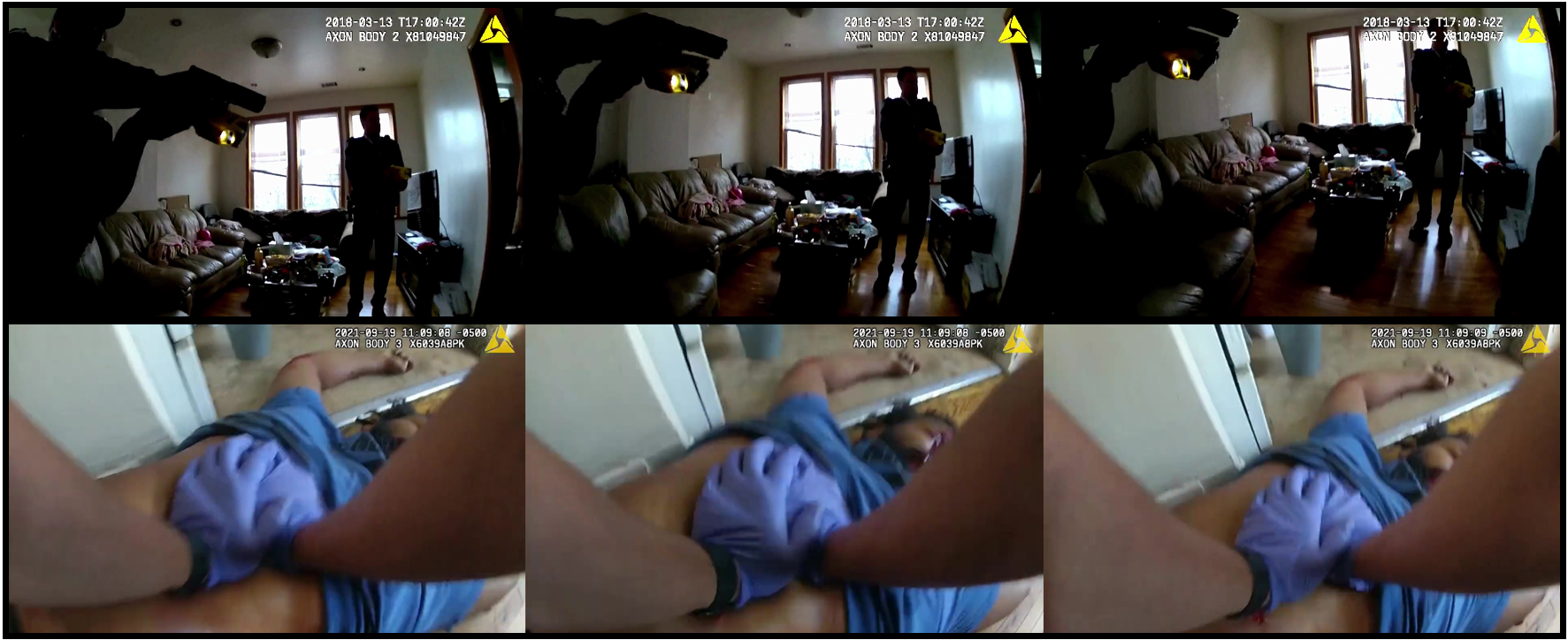}
        \caption{Misattribution}
        \label{minifig:misat}
    \end{subfigure}

    \caption{Examples of common failure modes. \datasetname{} combines in-the-wild egocentric video challenges with law-enforcement activity recognition. These include low-light occlusion, motion blur, rapid event occurrence, object misidentification, and action misattribution, which frequently co-occur within the same clip. Together, they highlight the gap between benchmark performance and the reliability required for real-world deployment.}
    \label{fig:qualitative}
\end{figure}

\noindent
\textbf{Qualitative analysis.} We observe consistent failure modes across all classification models. \Cref{fig:qualitative} groups representative examples by type, with each sequence showing three uniformly sampled frames from a one-second clip. 
Performance drops in low-light conditions, where flashlights create over- and underexposed regions that obscure actions such as ``Civilian-On-Ground'' or ``Any Officer-Handcuffing'' leading to false negatives (\cref{minifig:lowlight}). Blur from motion or lens obstructions also degrades performance, as in \cref{minifig:blur} (bottom), where a water droplet hides a gun. Short event durations cause additional errors (\cref{minifig:fasto}), since actions (civilian on ground or weapon out) only appear within part of a one-second window. We also observe false positives from misidentification (\cref{minifig:misid}), where objects like flashlights or red police tape are confused with firearms. In addition, models rely on spurious correlations, such as using a red T-shirt as a cue for injury (\cref{minifig:scorr}), and misattribute actions in complex scenes, such as assigning a taser held by another officer to the BWC wearer (\cref{minifig:misat}).

\section{Real-World Police Department Application}
\label{sec:real_world_pd}

\datasetname was designed by both social and computer scientists with the goal of supporting responsible analysis of real-world BWC footage. Many law enforcement agencies maintain large repositories of videos which capture police-civilian interactions of significant evidentiary value, yet the volume of this footage makes consistent human review practically infeasible. \datasetname provides a foundation for developing models that can efficiently identify events of interest.

In ongoing work, we have found that models trained on \datasetname can successfully support large-scale analysis of body-worn camera videos in a human-in-the-loop setting, making systematic review of large BWC repositories tractable. In this setting, we focus on the action class ``BWC Wearer-Physical Interaction.'' Rather than treating model predictions as final determinations, outputs are used to surface candidate video segments that are subsequently reviewed and verified by trained annotators. This workflow has proven effective at surfacing relevant events for review and illustrates the potential of models trained on \datasetname to support scalable analysis of body-worn camera footage.

This deployment suggests two preliminary insights. First, models trained on \datasetname learn representations that transfer to real-world BWC data. Second, the immediate value lies in assisting human reviewers by narrowing attention to potentially critical moments. This way, \datasetname facilitates scalable auditing and retrospective analysis of large volumes of body-worn camera footage. A full analysis of this deployment will be presented in a forthcoming companion paper. We include this discussion to emphasize that \datasetname is not only a challenging benchmark, but also a concrete example of how to design pipelines that can be deployed in real-world settings where human judgment remains central.

%% file: ethics.tex
\section{Ethical Concerns}
In this work, we study police body-worn camera (BWC) footage, a sensitive domain in which individuals may be recorded without consent. Although all videos in \datasetname are publicly available, privacy concerns remain. We restrict annotations to coarse action-level labels and avoid personal attribute annotations. We also recognize that BWC footage can be distressing for annotators and viewers. For annotators, we mitigate this through assigning narrowly scoped tasks for short clips (instead of full length videos), providing intensive training, and developing a custom tool designed for this type of video. For viewers, we provide a trigger warning urging caution before downloading and examining the dataset. 

We acknowledge that releasing BWC footage raises ethical questions, while recognizing that withholding it is not a normatively neutral choice. BWC technology exists to promote accountability, and the public has a role in that process. We believe it is appropriate to release such recordings when there is a clear public interest (\eg promoting accountability, enabling independent oversight, informing public understanding of policing and the lawful use of force, or supporting research) and appropriate safeguards protect the dignity of those depicted.

Beyond privacy, automated analysis of police footage carries broader risks, including reinforcing institutional biases and enabling unaccountable decision-making. High performance on \datasetname should not be interpreted as a basis for deployment. Instead we aim to highlight the potential and limitations of current systems. It is also important to note that the data may reflect demographic and institutional biases tied to the jurisdictions from which it was collected.

%% file: 7_conclusion.tex
\section{Conclusion}
\label{sec:conclusion}

In this work, we present \datasetname, a dataset of egocentric police–civilian interactions designed to challenge vision models with complex, real-world scenarios. Our analysis shows that police-recorded footage is a particularly difficult domain. Hence, \datasetname serves as a challenging benchmark for computer vision researchers focused on advancing research on vision models. Simultaneously, we believe that \datasetname can provide a resource for developing video understanding systems that are not only more capable, but also more responsible and accountable. As automated video analysis becomes more prevalent in law enforcement, the lack of transparent, rigorous benchmarks increases the risk of unexamined bias and system failures. We hope \datasetname encourages the community to confront the ethical, technical, and social challenges that arise as models are increasingly deployed in sensitive, real-world environments.

%% file: X_suppl.tex
{\large
\vspace{3em}
\noindent \hyperref[apx:limitations]{\textbf{A. Limitations}}

\quad \hyperref[apx:annotation_ambiguity]{\textbf{A.1 Annotation Ambiguity}}

\quad \hyperref[apx:limited_actions]{\textbf{A.2 Limited Actions}}

\quad \hyperref[apx:datasetbias]{\textbf{A.3 Dataset Bias Toward Firearms-Related Incidents}}

\noindent \hyperref[apx:add_related_works]{\textbf{B. Additional Related Works}}

\noindent \hyperref[apx:add_stats]{\textbf{C. Additional \datasetname{} Statistics}}

\quad \hyperref[apx:class_cooccurrence]{\textbf{C.1 Detailed Class Distribution Per Department}}

\quad \hyperref[apx:class_cooccurrence]{\textbf{C.2 Class Co-Occurrence}}

\quad \hyperref[apx:nonbwc]{\textbf{C.3 Non-BWC Footage Removal}}

\quad \hyperref[apx:iaa]{\textbf{C.4 Inter-Annotator Agreement}}

\noindent \hyperref[apx:exp_details]{\textbf{D. Experiment Details and Results}}

\quad \hyperref[apx:classification_eval]{\textbf{D.1 Classification Evaluation}}

\quad \hyperref[apx:mcq]{\textbf{D.2 MCQ}}

\noindent \hyperref[apx:ann_mm]{\textbf{E. Annotator Management}}

\quad \hyperref[apx:orig_strategy]{\textbf{E.1 Original Annotation Strategy}}

\quad \hyperref[apx:updated_strategy]{\textbf{E.2 Updated Annotation Strategy}}

\noindent \hyperref[apx:examples]{\textbf{F. Examples}}
}

\newpage

\section{Limitations}
\label{apx:limitations}

\subsection{Annotation Ambiguity}
\label{apx:annotation_ambiguity}
Annotators were instructed to label each second independently, based solely on what is visually observable, \emph{without} drawing inferences from surrounding footage. This approach can introduce ambiguities, \eg, a firearm that is clearly visible in second $k$ may become occluded or motion-blurred in $k+1$. A human observer with full video context could recognize it. However our protocol requires marking $k+1$ as negative for ``BWC Weapon Out''. We chose this approach to minimize bias, false positives and ensure consistency between annotators. However, we note that this might introduce false negatives at temporal boundaries.

\subsection{Limited Actions}
\label{apx:limited_actions}
\datasetname covers nine action classes. We selected these to balance social science relevance with feasibility for computer vision research. However, future work should consider broadening this taxonomy of policing behaviors to allow for more extended downstream use cases. We initially considered additional classes such as ``Civilian Weapon Out'' or ``Officer Pursuing Civilian on Foot''. Yet, they occur too infrequently in our corpus to provide sufficient positive examples for model training. This is a reflection of one of the challenges in high-stakes domains: the events most critical to document are often the rarest to observe. 

\subsection{Dataset Bias Toward Firearms-Related Incidents}
\label{apx:datasetbias}
\datasetname's primary data source for training and evaluation (COPA) releases recordings primarily for incidents involving civilian deaths, injuries, or officer firearm discharge, causing shooting-related incidents to be overrepresented in \datasetname. Models trained on this data may therefore transfer less well to other departments or incident types. This is consistent with the lower OOD performance observed on Pasadena footage.

\newpage
\section{Additional Related Works}
\label{apx:add_related_works}
\noindent
\textbf{Video models.} Early approaches to video understanding adapted standard 2D image architectures to the temporal domain by inflating their kernels into 3D~\cite{resnet3d,i3d,c3d}. Other works introduced vision-specific components such as temporal shifting~\cite{tsm}, dual-pathway processing~\cite{slowfast}, and video-specific transformer architectures~\cite{vivit,timesformer,mvitv1,mvitv2,uniformer}. The introduction of image embedding models~\cite{clip,dinov2,hiera,siglip2} has also motivated video-based embedding models like VideoMAEv2~\cite{videomaev2} and X-CLIP~\cite{xclip}.
More recently, there has been a shift towards Vision-Language Models (VLMs) that bridge pre-trained visual encoders with large language models. To manage extended video contexts, common architectures~\cite{Maaz2023VideoChatGPT,llavamini,li2024llamavid,lin2024video,merv,hong2024cogvlm2} employ a number of different mechanisms, ranging from spatio-temporal fusion~\cite{merv} to efficient temporal pooling~\cite{Maaz2023VideoChatGPT,li2024llamavid}. Current state-of-the-art open-source models~\cite{qwen25,internvl3,videollama3} achieve robust performance through training on vast-scale datasets combined with advanced alignment strategies. Proprietary models like Gemini-2.5~\cite{gemini25} natively accept video input, while GPT-4~\cite{gpt4} requires the workaround of feeding individual frames as images. \\

\section{Additional \datasetname{} Statistics}
\label{apx:add_stats}

\subsection{Detailed Class Distribution Per Department}

\begin{table*}[]
\centering
\captionsetup{justification=centering}
\begin{adjustbox}{width=\textwidth, valign=t}
\begin{tabular}{l|r|r|r|r|r|r|r||rr}
\hline
\multirow{2}{*}{Class} & \multicolumn{1}{c|}{COPA} & \multicolumn{1}{c|}{Pasadena} & \multicolumn{1}{c|}{D.C. PD} & \multicolumn{1}{c|}{Dallas PD} & \multicolumn{1}{c|}{Los Angeles PD} & \multicolumn{1}{c|}{Misc. PDs} & \multicolumn{1}{c||}{San Antonio PD} & \multicolumn{2}{c}{\datasetname\ ($\sim$185 hrs)} \\
 & \multicolumn{1}{c|}{\# seconds} & \multicolumn{1}{c|}{\# seconds} & \multicolumn{1}{c|}{\# seconds} & \multicolumn{1}{c|}{\# seconds} & \multicolumn{1}{c|}{\# seconds} & \multicolumn{1}{c|}{\# seconds} & \multicolumn{1}{c||}{\# seconds} & \# seconds & \% \\
\hline
BWC Physical Intervention & 13,772 & 1,749 & 5,119 & 1,631 & 705 & 9,520 & 61 & 32,557 & 4.89\% \\
BWC Medical Treatment & 6,387 & 851 & 270 & 36 & 14 & 308 & 16 & 7,882 & 1.18\% \\
BWC Weapon Out & 23,961 & 1,142 & 1,124 & 843 & 1,734 & 10,303 & 1,059 & 40,166 & 6.03\% \\
BWC Running & 2,893 & 434 & 534 & 90 & 38 & 2,267 & 368 & 6,624 & 0.99\% \\
Other Physical Intervention & 29,167 & 2,553 & 4,730 & 1,117 & 459 & 6,825 & 92 & 44,943 & 6.75\% \\
Other Medical Treatment & 15,810 & 1,376 & 418 & 61 & 20 & 543 & 0 & 18,228 & 2.74\% \\
Handcuffing & 1,138 & 152 & 719 & 243 & 98 & 1,187 & 63 & 3,600 & 0.54\% \\
Civilian Injured & 25,274 & 1,041 & 371 & 50 & 148 & 1,224 & 0 & 28,108 & 4.22\% \\
Civilian On Ground & 43,083 & 2,865 & 5,100 & 1,688 & 1,676 & 9,193 & 368 & 63,973 & 9.61\% \\
\hline
\textit{Any Class} & 75,542 & 6,122 & 10,614 & 3,523 & 3,322 & 29,838 & 1,778 & 130,739 & 19.63\% \\

\hline
\end{tabular}
\label{tab:summary_stats_at_second_level_detailed}
\end{adjustbox}
\end{table*}

\newpage
\subsection{Class Co-Occurrence}
\label{apx:class_cooccurrence}

\begin{figure}
    \vspace{-1em}
    \centering
    \includegraphics[width=0.9\linewidth]{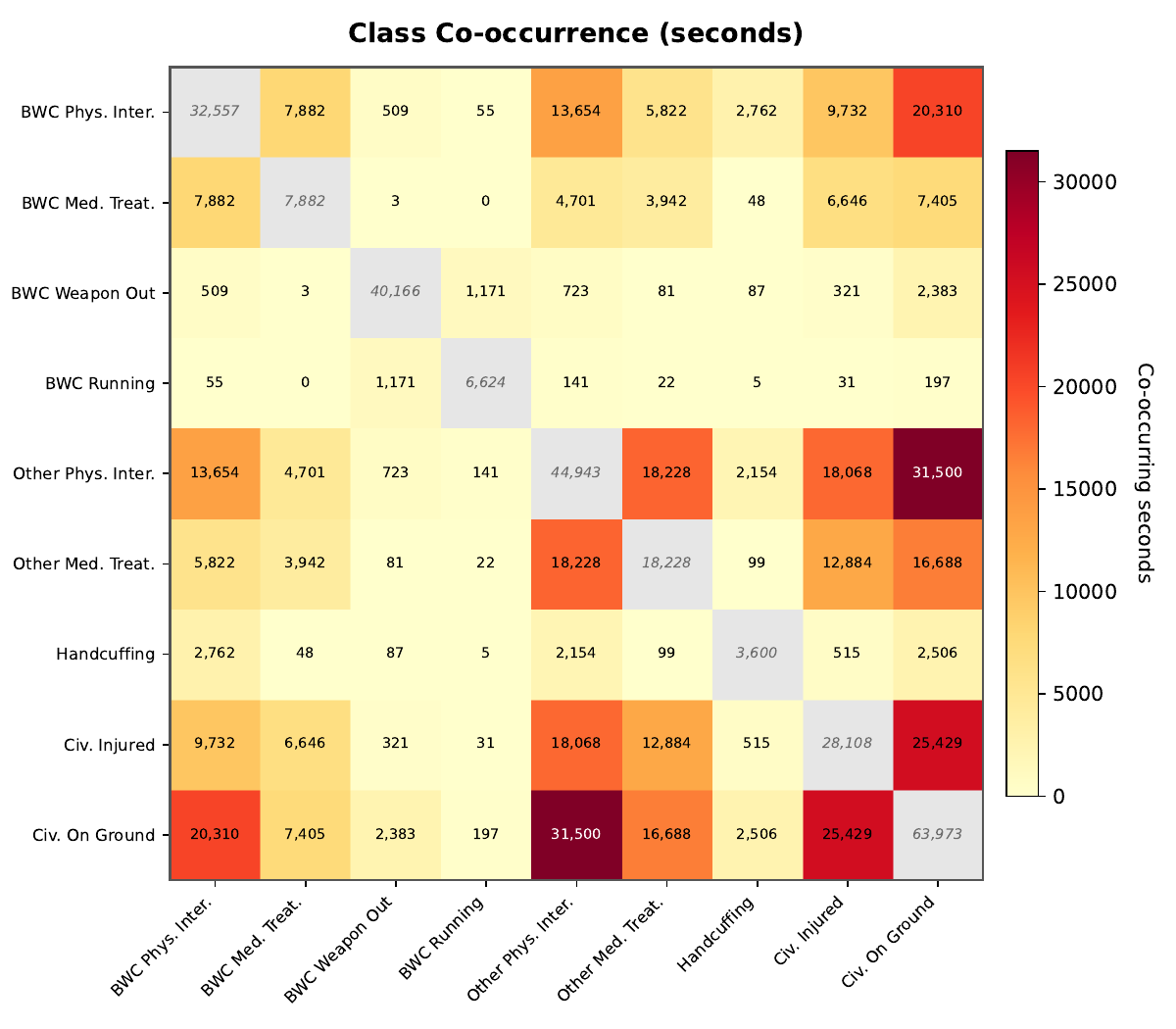}
    \label{fig:class_cooccurence}
    \vspace{-3em}
\end{figure}

\subsection{Non-BWC Footage Removal}
\label{apx:nonbwc}

\begin{figure}[!h]
    \vspace{-1em}
    \centering
    \includegraphics[width=\linewidth]{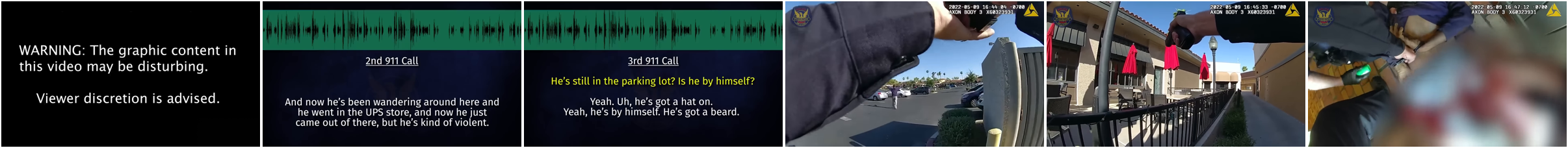}
    \begin{minipage}{0.19\linewidth}
        \centering
        \includegraphics[width=\linewidth]{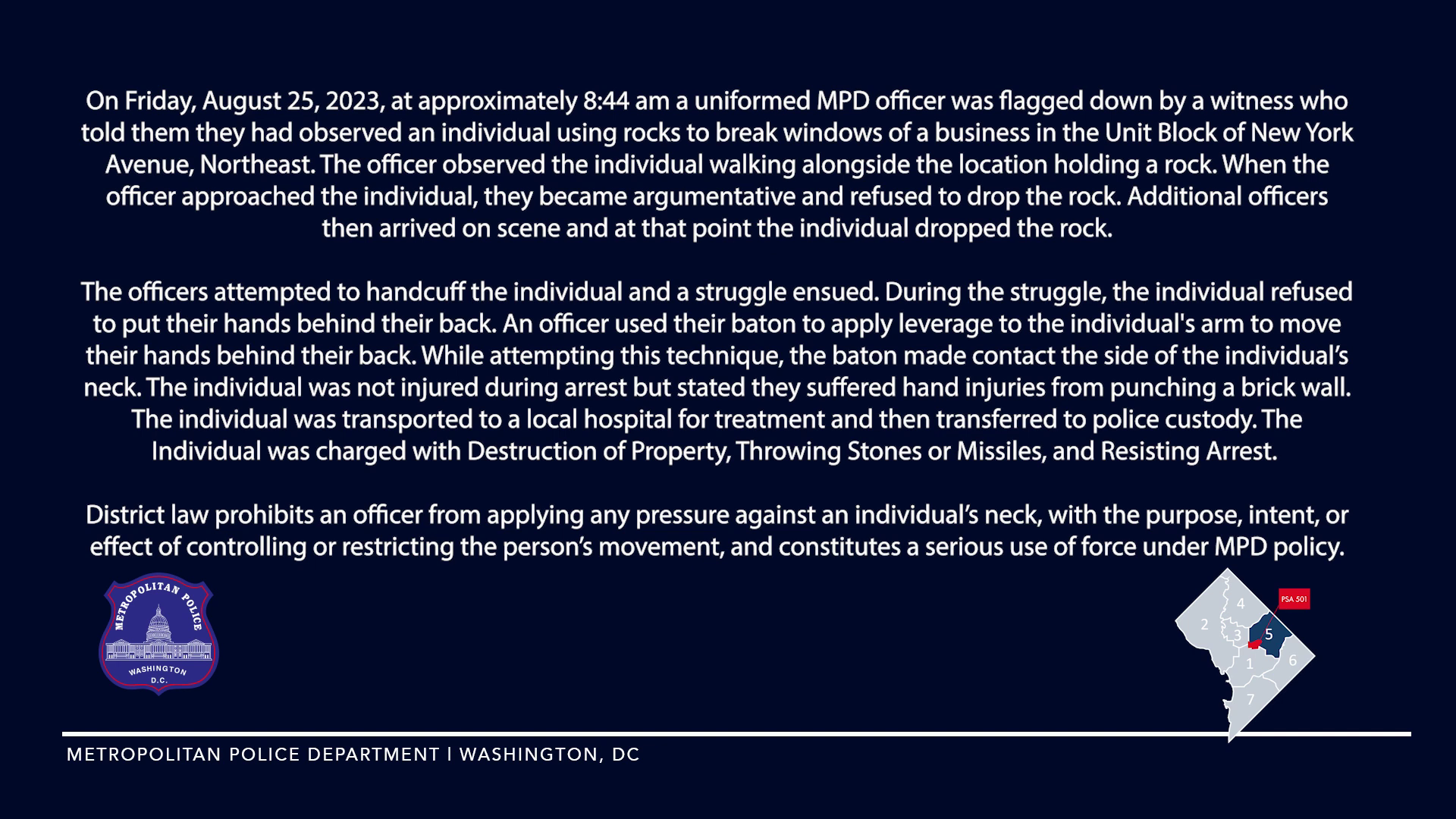}
    \end{minipage}
    \hfill
    \begin{minipage}{0.19\linewidth}
        \centering
        \includegraphics[width=\linewidth]{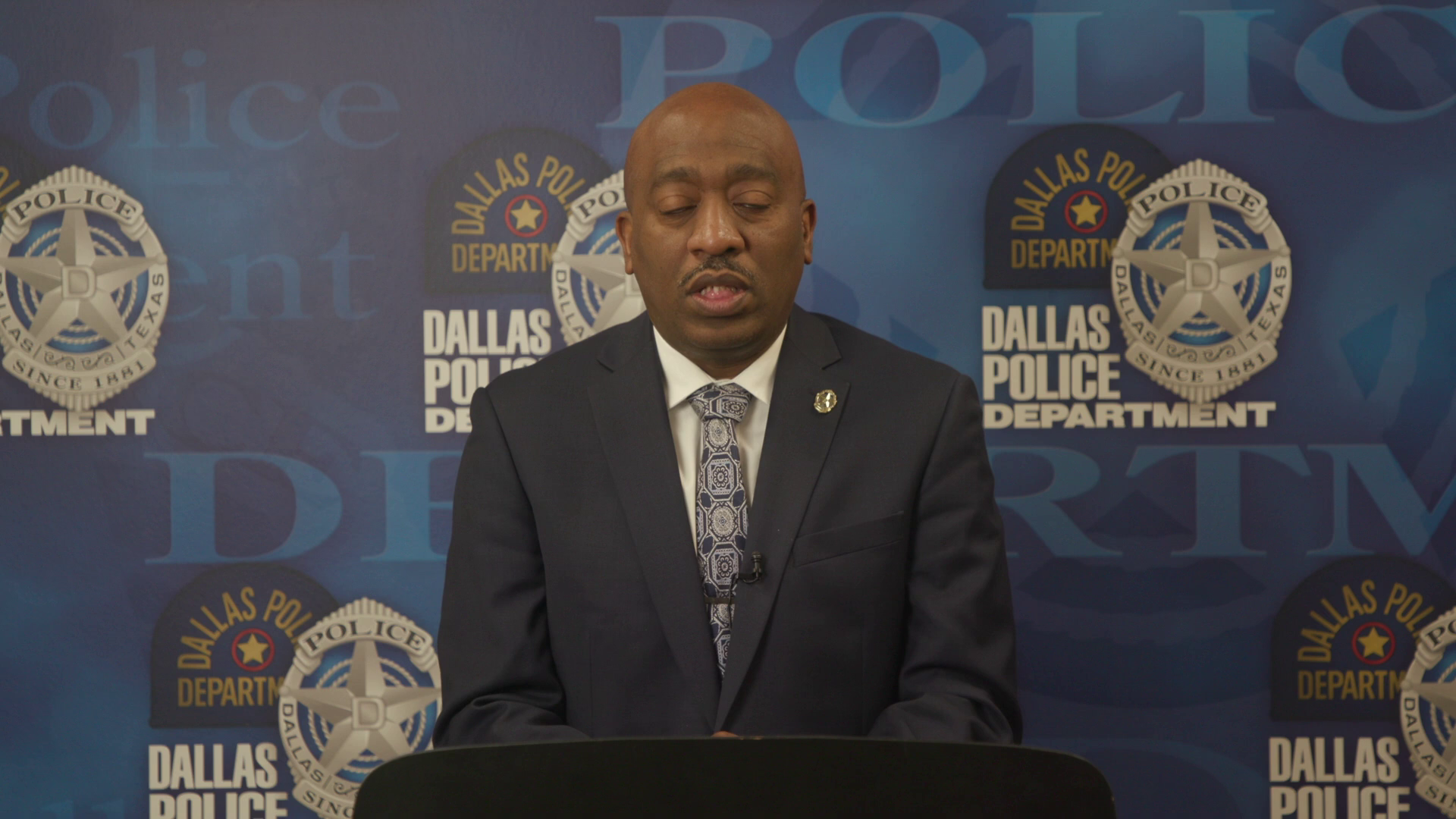}
    \end{minipage}
    \hfill
    \begin{minipage}{0.19\linewidth}
        \centering
        \includegraphics[width=\linewidth]{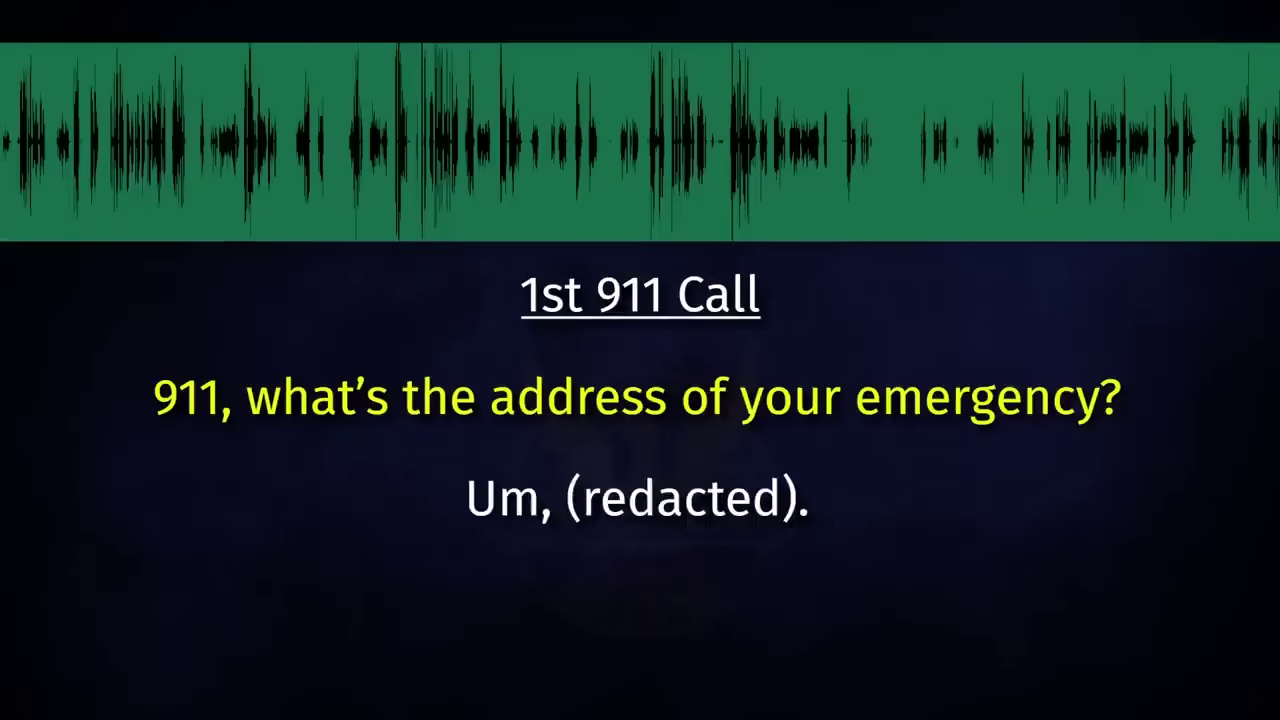}
    \end{minipage}
    \hfill
    \begin{minipage}{0.19\linewidth}
        \centering
        \includegraphics[width=\linewidth]{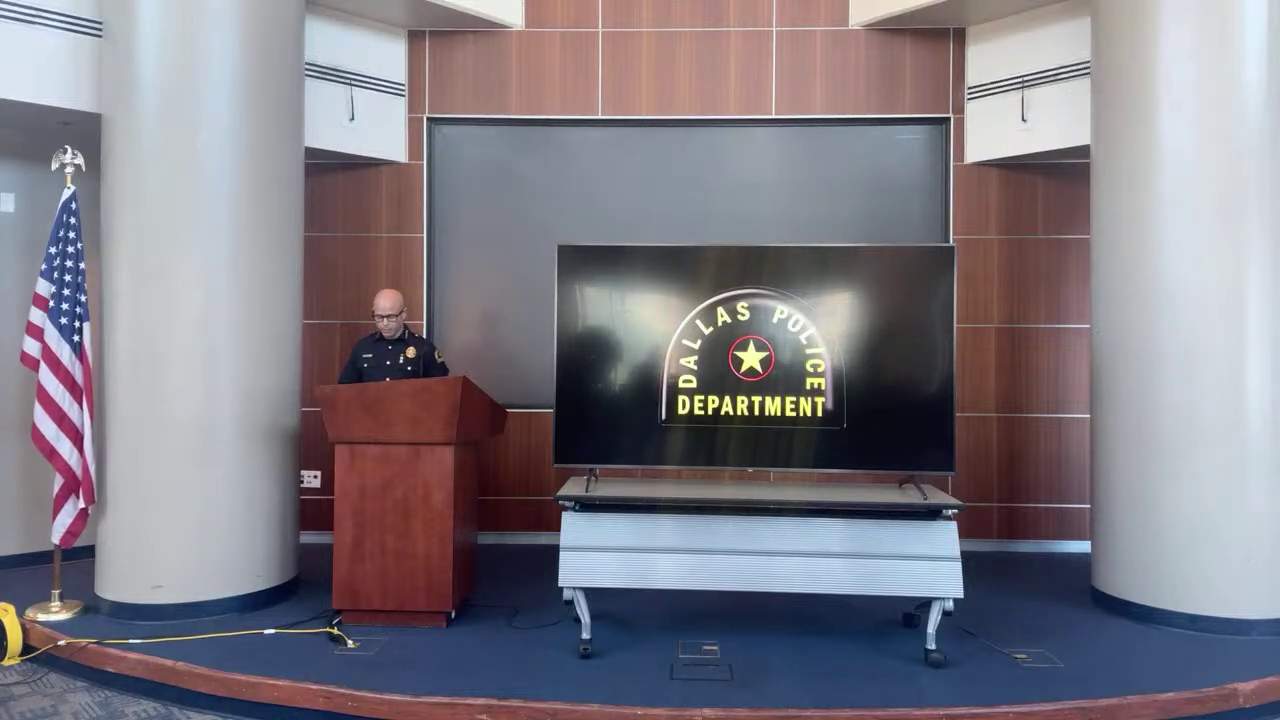}
    \end{minipage}
    \hfill
    \begin{minipage}{0.19\linewidth}
        \centering
        \includegraphics[width=\linewidth]{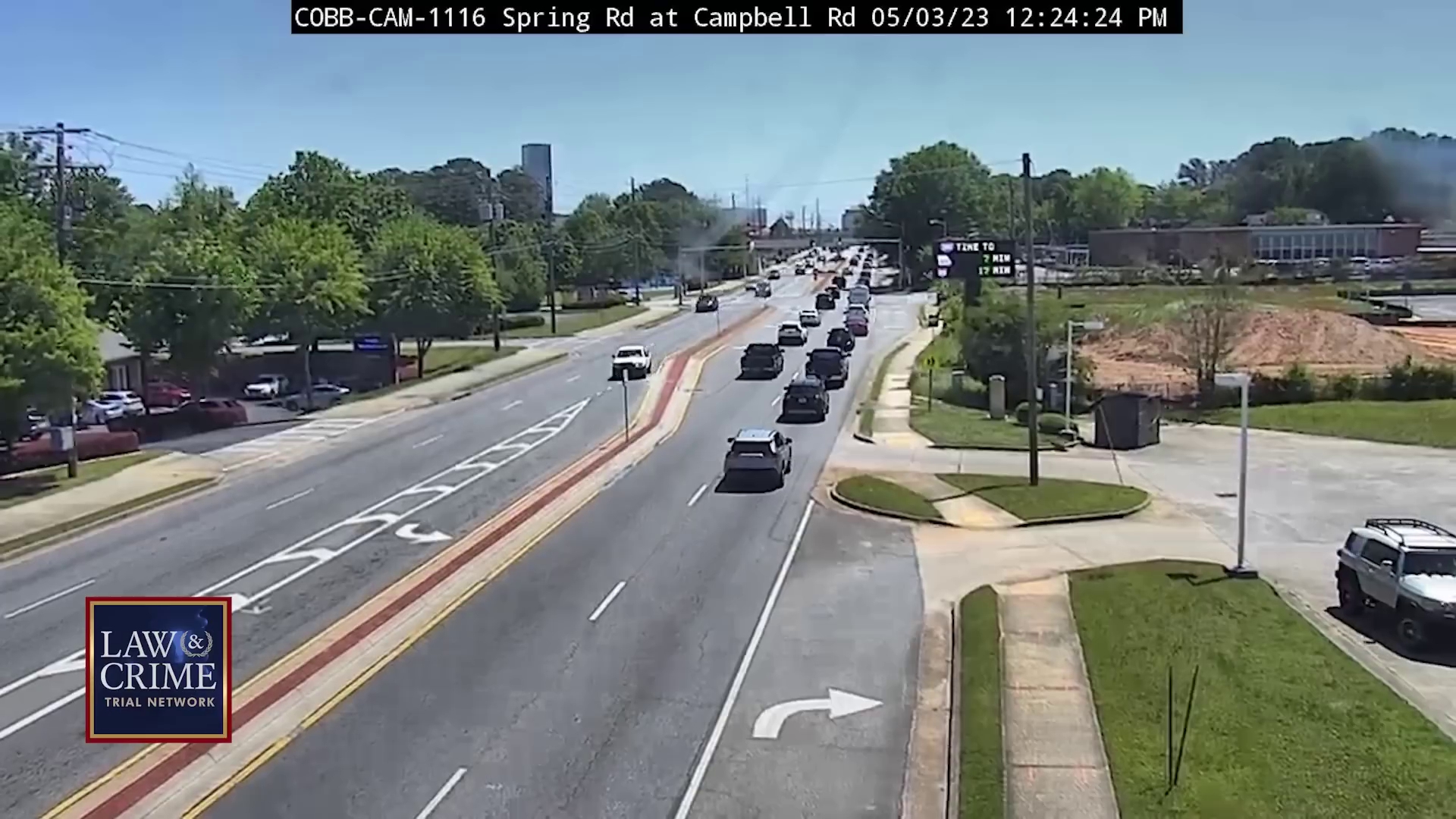}
    \end{minipage}
    \caption{Some videos contains non-BWC footage. \textbf{Top} Example video that starts with non-BWC footage before displaying an officer in action. \textbf{Bottom} Examples of non-BWC frames encountered in the police-released videos.}
    \label{fig:apx:output_figure}
\end{figure}

\noindent
\Cref{fig:apx:output_figure} illustrates that some videos contain segments that are not captured by body-worn cameras (BWC). All videos that we have collected originate from Axon devices and contain embedded overlays such as a timestamp, and the Axon logo. To filter out non-BWC footage, we apply an off-the-shelf OCR tool~\cite{jaidedai_easyocr} to detect these embedded text elements. Specifically, we drop all the frames that do not have \texttt{AXON BODY} written on the top right corner.

\newpage
\subsection{Inter-Annotator Agreement}
\label{apx:iaa}

\begin{table}[h]
    \centering
    \caption{Inter-annotator agreement (Krippendorff's alpha~\cite{krippendorff}) per action label.}
    \label{tab:iaa}
    \renewcommand{\arraystretch}{1.2}
    \resizebox{0.4\linewidth}{!}{
        \begin{tabular}{l l c}
            \toprule
            \textbf{Role} & \textbf{Action} & \textbf{IAA} \\
            \midrule
            \multirow{4}{*}{BWC Wearer}
                & Physical Interaction & 80.88 \\
                & Medical Treatment    & 86.52 \\
                & Weapon Out           & 84.95 \\
                & Running              & 82.68 \\
            \midrule
            \multirow{2}{*}{Other Officer}
                & Physical Interaction & 75.17 \\
                & Medical Treatment    & 64.00 \\
            \midrule
            Any Police Officer & Handcuffing & 71.99 \\
            \midrule
            \multirow{2}{*}{Civilian}
                & Visibly Injured & 80.74 \\
                & On Ground       & 87.45 \\
            \bottomrule
        \end{tabular}
        }
\end{table}

\section{Experiment Details and Results}
\label{apx:exp_details}
\subsection{Classification Evaluation}
\label{apx:classification_eval}
We follow the linear probing recipe proposed by \cite{dinov2}. Concretely we do a grid search over learning rates (1e-4, 2e-4, 5e-4, 1e-3, 2e-3, 5e-3, 1e-2, 2e-2, 5e-2, 0.1, 0.2, 0.5), layer depth (1, 4) and embedding representation (either mean pool frame level representation across all frames in a given second or concatenate them). We find the best combination on the validation set for each model and evaluate on the different test sets. The random baseline is a dummy model that always, for any given class in any given second, predicts ``1''. Recall is 100. 

\begin{table*}[h]
\centering
\caption{Classification results (mF1 and mAP) across all durations and splits.}
\label{tab:results_mf1_map}
\resizebox{\linewidth}{!}{
\begin{tabular}{lcccccccccccccccccc}
\toprule
 & \multicolumn{6}{c}{1s} & \multicolumn{6}{c}{10s} & \multicolumn{6}{c}{1min} \\
\cmidrule(lr){2-7} \cmidrule(lr){8-13} \cmidrule(lr){14-19}
 & \multicolumn{2}{c}{ID} & \multicolumn{2}{c}{OOD-T} & \multicolumn{2}{c}{OOD-L} & \multicolumn{2}{c}{ID} & \multicolumn{2}{c}{OOD-T} & \multicolumn{2}{c}{OOD-L} & \multicolumn{2}{c}{ID} & \multicolumn{2}{c}{OOD-T} & \multicolumn{2}{c}{OOD-L} \\
\cmidrule(lr){2-3} \cmidrule(lr){4-5} \cmidrule(lr){6-7} \cmidrule(lr){8-9} \cmidrule(lr){10-11} \cmidrule(lr){12-13} \cmidrule(lr){14-15} \cmidrule(lr){16-17} \cmidrule(lr){18-19}
Model & mF1 & mAP & mF1 & mAP & mF1 & mAP & mF1 & mAP & mF1 & mAP & mF1 & mAP & mF1 & mAP & mF1 & mAP & mF1 & mAP \\
\midrule
\textit{Random Baseline} & 7.91 & 4.21 & 13.93 & 7.81 & 6.66 & 3.49 & 11.28 & 6.15 & 20.18 & 11.84 & 8.44 & 4.47 & 18.14 & 10.35 & 32.27 & 20.49 & 13.32 & 7.27 \\
CLIP \cite{clip} & \underline{39.15} & \textbf{47.92} & 42.08 & \underline{55.81} & 28.52 & 42.29 & 48.54 & \underline{55.27} & 49.78 & 62.63 & 40.62 & 50.22 & 53.53 & 61.52 & 54.92 & 71.63 & 46.11 & 58.95 \\
DINOv2 \cite{dinov2} & \textbf{40.22} & \underline{47.42} & \underline{43.90} & 54.03 & 25.68 & 39.62 & \underline{49.35} & 54.88 & \underline{50.53} & 61.54 & 35.31 & 47.23 & 54.19 & 61.44 & \underline{56.93} & 70.51 & 42.03 & 55.31 \\
Hiera \cite{hiera} & 34.22 & 38.88 & 40.86 & 50.78 & 26.08 & 27.63 & 44.80 & 47.24 & 48.86 & 57.59 & 33.35 & 35.12 & 48.79 & 54.68 & 56.43 & 66.88 & 35.58 & 43.75 \\
X-CLIP \cite{xclip} & 36.24 & 46.40 & 39.86 & 54.96 & \underline{30.84} & \textbf{46.01} & 49.30 & 54.56 & 48.32 & \underline{63.01} & \underline{45.84} & \textbf{54.22} & \underline{55.40} & \underline{61.86} & 54.67 & \underline{72.07} & \textbf{51.78} & \textbf{63.32} \\
VideoMAE v2 \cite{videomaev2} & 37.91 & 46.46 & \textbf{48.50} & \textbf{59.76} & \textbf{30.92} & \underline{44.14} & \textbf{51.79} & \textbf{55.46} & \textbf{59.27} & \textbf{67.41} & \textbf{46.15} & \underline{52.44} & \textbf{58.37} & \textbf{63.75} & \textbf{63.72} & \textbf{74.22} & \underline{49.78} & \underline{59.62} \\
\bottomrule
\end{tabular}
}
\end{table*}

\input{supp_class_result_detailed}

\newpage
\subsection{MCQ}
\label{apx:mcq}

\subsubsection{Prompt}

We use following prompt for our MCQ task. 
\begin{center}
\fbox{
    \begin{minipage}{0.9\linewidth}
        {\ttfamily
        Which action is happening in this video clip? Select the correct answer from the following options. Write your answer as only one of A, B, C, D, E and nothing else.\\

        A. The officer wearing the camera is providing visible medical assistance.

        B. The officer wearing the camera is running.

        C. An officer is visibly using or attempting to use handcuffs.

        D. A civilian displays visible wounds or is bleeding.

        E. None of the above
        }
    \end{minipage}
}
\end{center}

For each of the action labels, we use following descriptions:
\begin{itemize}
    \item \texttt{BWC-Physical Interaction}: The officer wearing the camera is visibly touching a civilian.
    \item \texttt{BWC-Medical Treatment}: The officer wearing the camera is providing visible medical assistance.
    \item \texttt{BWC-Running}: The officer wearing the camera is running.
    \item \texttt{BWC-Weapon Out}: A firearm, a taser, a gun or a rifle of the officer wearing the camera is visible in the video.
    \item \texttt{Civilian-Injured}: A civilian displays visible wounds or is bleeding.
    \item \texttt{Other Officer-Physical Interaction}: An officer (not the one recording) is visibly touching a civilian.
    \item \texttt{Other Officer-Medical Treatment}: An officer (not the one recording) is providing visible medical assistance.
    \item \texttt{Any Officer-On Ground}: An officer is visibly using or attempting to use handcuffs.
    \item \texttt{Civilian-On Ground}: A civilian is sitting, kneeling, or lying on the ground.
\end{itemize}

\subsubsection{Parsing}

Despite giving clear instructions to only output one of five letters, VLMs still show variation in their outputs, \eg, \textcolor{red}{\texttt{A}}, \textcolor{red}{\texttt{A.}}, \textcolor{red}{\texttt{(A)}}, \textcolor{red}{\texttt{a.}}, etc. To parse the output into single choice, we decapitalize the output and find the first occurrence of a, b, c, d, or e. Specifically, our parsing code looks like:
\begin{center}
    \begin{minipage}{0.99\linewidth}
        \begin{lstlisting}[language=Python,frame=single]
#prediction = "E) None of the above"
prediction = prediction.lower()
prediction = "".join([c for c in prediction if c in "abcde"])
if len(prediction) == 0:
    prediction = "F"
pred_idx = ord(prediction[0]) - ord("a")
        \end{lstlisting}
    \end{minipage}
\end{center}

\noindent
We observe a similar probability for each letter to be the answer, except for \texttt{E. None of the above}. We see some VLMs fail to follow the instructions and output video descriptions instead of outputting the corresponding letter (Format Error). With the description often starting with \texttt{The}, our parser assigns \texttt{E} as the model output, leading to accuracy lower than random guess.
To disentangle format error from the model performance, we use following regular expression \verb|r'^[^A-Za-z]*[A-E](?![A-Za-z])'|. The expression treats the output as ``Legal'' if the first alphabetical letter is from \texttt{A} to \texttt{E} and is not followed by another letter, \ie accepting \textcolor{red}{\texttt{A}}, \textcolor{red}{\texttt{A.}}, \textcolor{red}{\texttt{(A)}}, or \textcolor{red}{\texttt{A) An officer is...}} but not \textcolor{red}{\texttt{The video shows...}} or \textcolor{red}{\texttt{Answer is A}}. \Cref{tab:apx:pe} shows that LLaVA-Mini-8B~\cite{llavamini} produces many format errors, while other models show minimal to no parsing issues. 

\vspace{-1.5em}
\begin{table*}[h]
\centering
\caption{Parsing Error across VLMs.}
\vspace{-1em}
\label{tab:apx:pe}
\resizebox{0.45\linewidth}{!}{
\begin{tabular}{lrr}
\toprule
Model & Overall Acc. & \# of Parsing Error \\
\midrule
LLaVA-Mini-8B~\cite{llavamini}          & 15.4 & 11998 \\
GPT-4.1-Nano~\cite{gpt4}                & 17.3 & 40    \\
LLaMA-VID-7B~\cite{li2024llamavid}      & 36.6 & 0     \\
VideoLLaVA-7B~\cite{lin2024video}       & 36.6 & 0     \\
MERV-7B~\cite{merv}                     & 41.7 & 0     \\
CogVLM2-Video-8B~\cite{hong2024cogvlm2} & 46.9 & 0     \\
Qwen2.5-VL-7B~\cite{qwen25}              & 54.4 & 0     \\
VideoLLaMA3-7B~\cite{videollama3}       & 59.4 & 0     \\
InternVL3-8B~\cite{internvl3}           & 63.9 & 0     \\
Gemini 2.5 Flash~\cite{gemini25}        & 67.5 & 1064  \\
GPT-4.1~\cite{gpt4}                     & 76.5 & 0     \\
Gemini 2.5 Pro~\cite{gemini25}          & 77.2 & 0     \\
\bottomrule
\end{tabular}
}
\end{table*}
\vspace{-3em}

\newpage
\subsubsection{Per class accuracy}
\; \\
\vspace{-1em}

\begin{table*}[h]
\centering
\caption{Per class accuracy on MCQ.}
\label{tab:apx:mcq_per_class}
\resizebox{\linewidth}{!}{

\begin{tabular}{c|c|c|c|c|c|c|c|c|c}
\toprule
\multicolumn{10}{c}{1s} \\
\midrule
Model & BWC Officer & BWC Officer & BWC Officer & BWC Officer & Civilian & Civilian & Any Officer & Other Officer & Other Officer \\
                                                & Medical Treatment & Physical Interaction & Running & Weapon Out & Injured & on Ground & Handcuffing & Medical Treatment & Physical Interaction \\
\midrule
\; LLaVA-Mini-8B~\cite{llavamini}          & 10.06             & 18.34                & 11.26             & 11.23             & 11.76             & 18.24             & 17.63             & 11.24             & 14.11                \\
\; GPT-4.1-Nano~\cite{gpt4}                & 5.20              & 10.74                & 0.79              & 5.86              & 9.81              & 46.15             & 24.31             & 13.82             & 20.31                \\
\; VideoLLaVA-7B~\cite{lin2024video}       & 48.79             & 31.06                & 30.70             & 42.50             & 58.17             & 52.70             & 29.76             & 44.08             & 19.31                \\
\; LLaMA-VID-7B~\cite{li2024llamavid}      & 44.95             & 25.46                & 65.25             & 49.39             & 41.58             & 30.26             & 31.62             & 42.31             & 28.17                \\
\; MERV-7B~\cite{merv}                     & 50.90             & 48.27                & 53.80             & 48.47             & 59.40             & 49.94             & 32.71             & 37.07             & 21.59                \\
\; CogVLM2-Video-8B~\cite{hong2024cogvlm2} & 44.31             & 49.50                & 38.33             & 66.88             & 58.01             & 34.09             & 61.20             & 39.95             & 37.58                \\
\; Qwen2.5-VL-7B~\cite{qwen25}              & 36.50             & 60.14                & 41.93             & 53.89             & 68.03             & 69.20             & 54.77             & 58.17             & 55.37                \\
\; VideoLLaMA3-7B~\cite{videollama3}       & 56.28             & 57.28                & 33.28             & 65.62             & 67.10             & 75.06             & 58.06             & 61.60             & 61.05                \\
\; InternVL3-8B~\cite{internvl3}           & 53.65             & 67.45                & 60.98             & 71.52             & 74.45             & 67.16             & 66.92             & 68.51             & 65.04                \\
\; Gemini 2.5 Flash~\cite{gemini25}        & 61.06             & \underline{70.49}    & 45.50             & 80.28             & 77.26             & 80.05             & 74.16             & 62.72             & 71.89                \\
\; Gemini 2.5 Pro~\cite{gemini25}          & \textbf{70.51}    & 70.37                & \underline{82.01} & \underline{81.13} & \textbf{85.00}    & \textbf{90.25}    & \underline{78.89} & \underline{75.84} & \textbf{76.30}       \\
\; GPT-4.1~\cite{gpt4}                     & \underline{69.44} & \textbf{77.11}       & \textbf{86.79}    & \textbf{84.45}    & \underline{80.99} & \underline{86.01} & \textbf{79.26}    & \textbf{76.97}    & \underline{74.85}    \\
\bottomrule
\multicolumn{10}{c}{10s} \\
\toprule
\; LLaVA-Mini-8B~\cite{llavamini}          & 12.39             & 18.23                & 12.83             & 10.40             & 11.84             & 12.29             & 22.63             & 10.71             & 14.60                \\
\; GPT-4.1-Nano~\cite{gpt4}                & 5.32              & 13.57                & 0.40              & 6.78              & 13.31             & 45.04             & 20.98             & 13.90             & 21.27                \\
\; VideoLLaVA-7B~\cite{lin2024video}       & 47.04             & 24.73                & 30.09             & 35.94             & 50.20             & 47.80             & 27.14             & 41.43             & 19.06                \\
\; LLaMA-VID-7B~\cite{li2024llamavid}      & 53.14             & 26.51                & 46.62             & 46.84             & 41.78             & 27.46             & 31.03             & 44.37             & 31.43                \\
\; MERV-7B~\cite{merv}                     & 51.48             & 40.09                & 50.84             & 44.48             & 54.59             & 49.81             & 30.72             & 40.53             & 26.16                \\
\; CogVLM2-Video-8B~\cite{hong2024cogvlm2} & 46.34             & 50.09                & 45.31             & 66.54             & 55.21             & 35.86             & 60.39             & 48.09             & 37.70                \\
\; Qwen2.5-VL-7B~\cite{qwen25}              & 47.41             & 65.19                & 62.74             & 52.60             & 70.38             & 72.22             & 52.37             & 57.77             & 54.01                \\
\; VideoLLaMA3-7B~\cite{videollama3}       & 62.64             & 58.65                & 47.95             & 66.91             & 60.71             & 74.26             & 60.12             & 67.14             & 59.20                \\
\; InternVL3-8B~\cite{internvl3}           & 59.63             & 65.76                & 67.71             & 64.52             & 71.71             & 70.15             & 59.86             & 71.87             & 62.51                \\
\; Gemini 2.5 Flash~\cite{gemini25}        & 61.28             & 70.48                & 75.32             & \underline{76.92} & 78.45             & 84.30             & 73.30             & 68.07             & 71.47                \\
\; GPT-4.1~\cite{gpt4}                     & \underline{72.91} & \textbf{77.18}       & \underline{82.10} & \textbf{77.96}    & \underline{79.72} & \underline{85.39} & \underline{77.38} & \textbf{79.04}    & \underline{72.71}    \\
\; Gemini 2.5 Pro~\cite{gemini25}          & \textbf{76.58}    & \underline{74.72}    & \textbf{83.88}    & 76.62             & \textbf{81.39}    & \textbf{88.70}    & \textbf{79.13}    & \underline{78.85} & \textbf{75.15}       \\
\bottomrule
\multicolumn{10}{c}{1min} \\
\toprule
\; GPT-4.1-Nano~\cite{gpt4}                & 3.08              & 2.54                 & 1.00              & 0.00              & 0.00              & 9.87              & 8.98              & 2.70              & 9.82                 \\
\; LLaVA-Mini-8B~\cite{llavamini}          & 8.37              & 13.69                & 7.41              & 9.39              & 10.08             & 15.25             & 21.22             & 6.67              & 16.33                \\
\; LLaMA-VID-7B~\cite{li2024llamavid}      & 48.68             & 9.22                 & 22.54             & 43.78             & 23.79             & 24.29             & 26.21             & 37.30             & 29.04                \\
\; VideoLLaVA-7B~\cite{lin2024video}       & 41.88             & 25.57                & 25.64             & 35.97             & 37.01             & 48.37             & 21.19             & 41.92             & 19.67                \\
\; MERV-7B~\cite{merv}                     & 54.12             & 39.92                & 45.05             & 42.03             & 53.87             & 45.07             & 25.63             & 42.51             & 27.27                \\
\; CogVLM2-Video-8B~\cite{hong2024cogvlm2} & 50.83             & 47.65                & 27.82             & 54.95             & 53.87             & 18.83             & 46.97             & 46.94             & 39.69                \\
\; Qwen2.5-VL-7B~\cite{qwen25}              & 61.35             & 63.07                & 47.76             & 51.39             & 68.18             & 72.45             & 51.75             & 58.15             & 57.67                \\
\; VideoLLaMA3-7B~\cite{videollama3}       & 63.09             & 59.29                & 25.32             & 63.92             & 63.01             & 80.12             & 53.03             & 64.90             & 59.13                \\
\; InternVL3-8B~\cite{internvl3}           & 64.77             & 68.01                & 49.53             & 60.70             & 66.88             & 69.51             & 46.81             & 68.21             & 63.13                \\
\; Gemini 2.5 Flash~\cite{gemini25}        & 67.35             & 70.11                & 62.98             & 76.34             & 76.38             & 82.15             & 67.87             & 66.67             & 65.85                \\
\; GPT-4.1~\cite{gpt4}                     & \underline{75.37} & \textbf{78.92}       & \underline{65.40} & \textbf{80.94}    & \underline{80.80} & \underline{86.80} & \underline{74.94} & \textbf{82.88}    & \textbf{68.84}       \\
\; Gemini 2.5 Pro~\cite{gemini25}          & \textbf{80.73}    & \underline{78.21}    & \textbf{71.11}    & \underline{78.80} & \textbf{81.61}    & \textbf{90.13}    & \textbf{76.43}    & \underline{80.30} & \underline{68.44}    \\

\bottomrule
\end{tabular}
}
\end{table*}

\section{Annotator Management}
\label{apx:ann_mm}

Given the uniqueness of our dataset, we consider it important to share key insights and lessons learned from our annotation process, including the challenges and limitations we encountered. As AI is increasingly applied to complex real-world settings, we hope these experiences will guide future researchers and stimulate discussion around best practices for high-quality annotation.

\subsection{Original Annotation Strategy}
\label{apx:orig_strategy}

Initially, we pursued an annotation strategy where, in the first stage, annotators labeled interactions between civilians and the body-camera wearer (BCW) as ``explicit'' if the civilian appeared in frame, or ``implicit'' if the interaction occurred off-frame but could reasonable be inferred to be occurring. This distinction was intended to capture interactions that were not directly visible. In Stage 2, annotators labeled approximately 50 classes across three subjects, BCW, other officers, and civilians, focusing on segments identified in Stage 1. Before beginning either stage, annotators received in-person and remote training, completed test annotations, and received detailed feedback to ensure high-quality labeling.

\subsection{Updated Annotation Strategy}
\label{apx:updated_strategy}

We found that asking annotators to review long video sequences, or to annotate implicit interactions over extended periods, led to fatigue, missed labels, and reduced annotation quality. To improve accuracy and reliability, we implemented several updates:

\begin{itemize}
\item \textbf{Reduced the number of classes:} Many initial classes appeared in only a few frames across nearly 200 hours of footage. We reduced the set from over 50 to 9 of the most frequent, important, and feasible classes, ensuring the resulting labels would be useful for model training.
\item \textbf{Dropped implicit/explicit distinctions:} We first told annotators to annotate things that they could reasonably infer were happening, but where outside the field of view as ``implicit'' and annotate things in the field of view as ``explicit''. However, we found that annotators struggled with this distinction. To eliminate ambiguity, we required annotators to label only actions with clear visual cues visible in the field of view of the camera during any one-second clip.
\item \textbf{Clarified class definitions:} We created a dedicated website with detailed definitions, including positive and negative examples for each class. Annotators completed a tutorial of 25 example frames, classifying each as positive or negative before beginning the main task.
\item \textbf{Shorter video portions:} Annotators were assigned short clips of up to 150 seconds (instead of longer videos) to reduce fatigue and maintain focus.
\item \textbf{One class at a time:} Annotators focused on labeling one class per session, which reduced cognitive load and improved accuracy for each label.
\item \textbf{Random audits:} Approximately 25\% of annotations were randomly reviewed by the project manager. Annotators updated any errors, improving overall label quality and providing iterative learning opportunities.
\end{itemize}

\newpage
\section{Examples EgoPolice: Randomly Sampled Frames}
\label{apx:examples}

\begin{figure*}[]
    \centering
    \begin{adjustbox}{width=0.8\textwidth}
    \includegraphics[width=\linewidth]{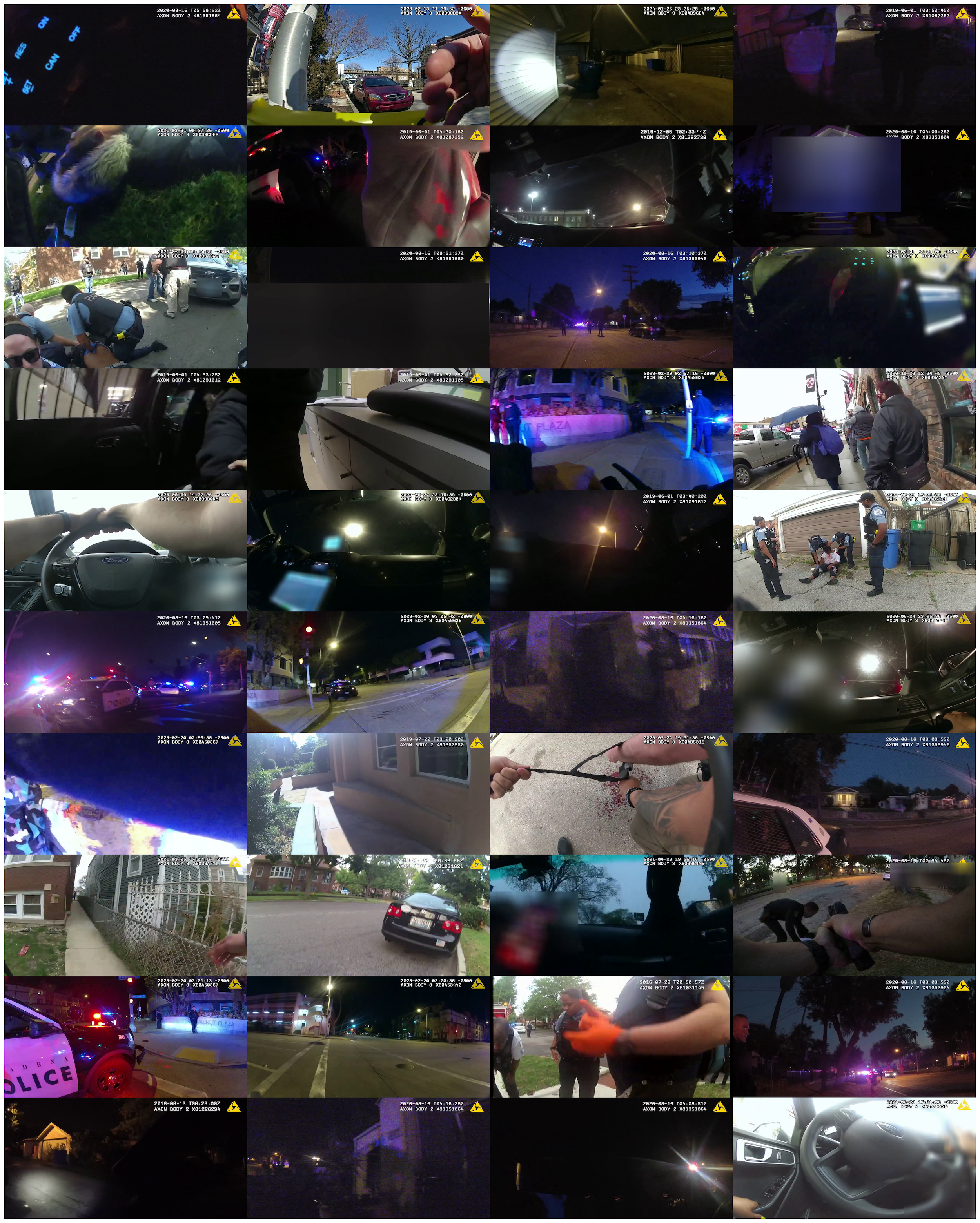}
    \end{adjustbox}
    \caption{Randomly Sampled Frames from COPA and Pasadena. Videos from COPA and Pasadena are minimally edited and best reflect real-world BWC footage.}
    \label{fig:copa_pasadena_examples}
\end{figure*}

\begin{figure*}[]
    \centering
    \begin{adjustbox}{width=0.8\textwidth}
    \includegraphics[width=\linewidth]{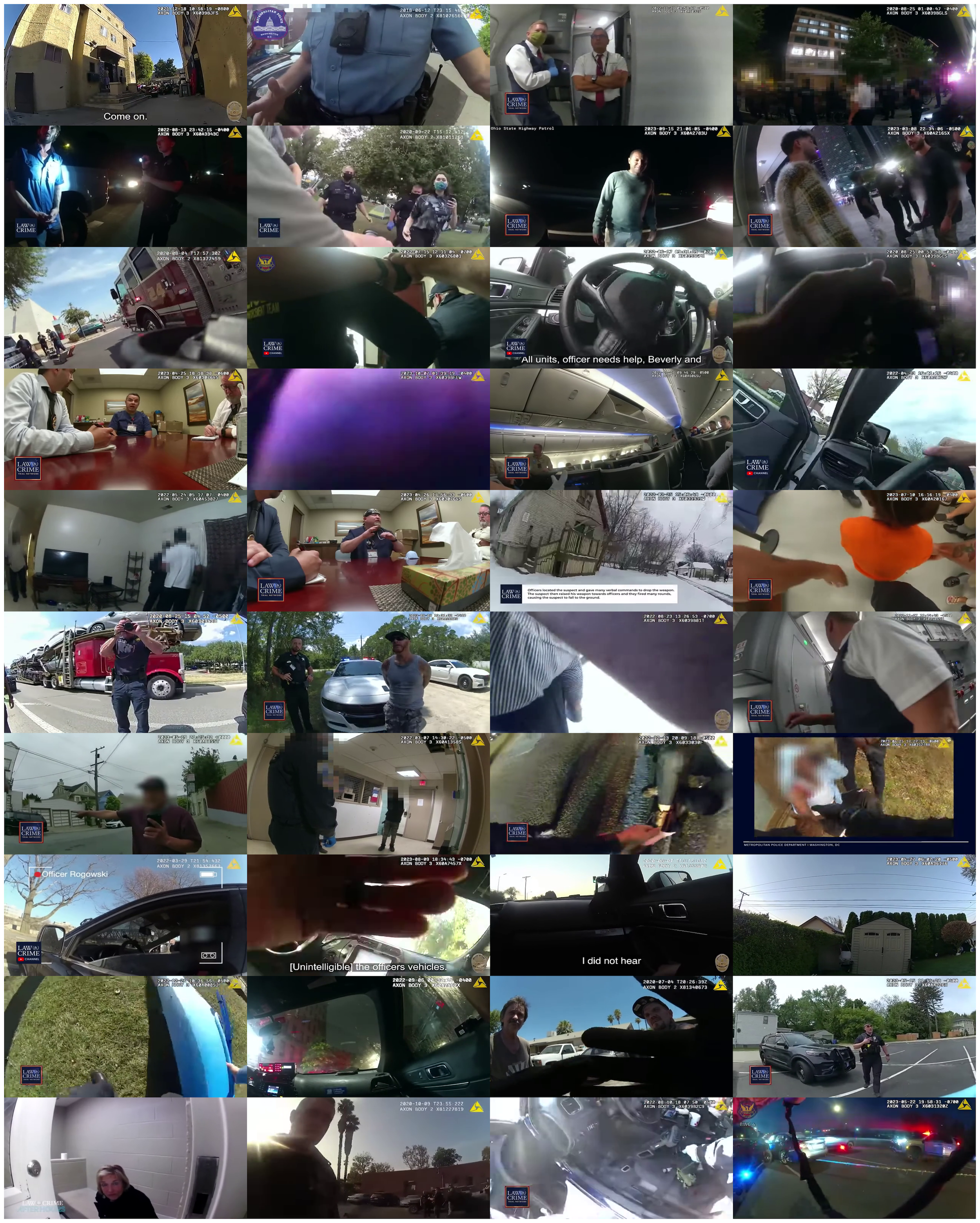}
    \end{adjustbox}
    \caption{Randomly Sampled Frames from departments other than COPA and Padadena. Videos from these sources are often edited (\eg, with overlaid subtitles) and are therefore treated as auxiliary training data. These videos are not included in validation or test splits.}
    \label{fig:other_department_examples}
\end{figure*}

%% file: supp_class_result_detailed.tex
\clearpage
\subsection{Per-Class Results: ID Split, 1s Clips}
\label{supp:per_class_id_1s}

\begin{center}
\captionof{table}{Per-class F1, Precision (P), Recall (R), AP: all models (ID, 1s).}
\label{tab:per_class_id_1s}

\begin{adjustbox}{width=0.75\linewidth}
\begin{tabular}{lcccccccccccc}
\toprule
 & \multicolumn{4}{c}{\textit{BWC Medical Treatment}} & \multicolumn{4}{c}{\textit{BWC Physical Interaction}} & \multicolumn{4}{c}{\textit{BWC Running}} \\
\cmidrule(lr){2-5} \cmidrule(lr){6-9} \cmidrule(lr){10-13}
Model & F1 & P & R & AP & F1 & P & R & AP & F1 & P & R & AP \\
\midrule
\textit{Random Baseline} & 2.90 & 1.47 & \textbf{100.00} & 1.47 & 5.91 & 3.04 & \textbf{100.00} & 3.04 & 0.88 & 0.44 & \textbf{100.00} & 0.44 \\
CLIP \cite{clip} & \underline{39.96} & 61.11 & 29.68 & \underline{55.14} & 63.38 & \textbf{70.32} & 57.68 & 66.11 & 2.86 & \textbf{20.34} & 1.54 & 6.83 \\
DINOv2 \cite{dinov2} & \textbf{42.40} & \underline{68.34} & \underline{30.74} & \textbf{58.41} & \textbf{67.30} & 69.15 & \underline{65.56} & \textbf{69.92} & 0.95 & 6.35 & 0.51 & 3.89 \\
Hiera \cite{hiera} & 35.81 & 60.25 & 25.47 & 48.13 & \underline{64.93} & 68.47 & 61.74 & 66.59 & 5.93 & 13.75 & 3.78 & 6.52 \\
X-CLIP \cite{xclip} & 30.80 & 63.59 & 20.33 & 52.38 & 62.95 & \underline{69.87} & 57.28 & 66.68 & \underline{7.05} & \underline{19.82} & 4.29 & \underline{7.81} \\
VideoMAE v2 \cite{videomaev2} & 31.16 & \textbf{71.67} & 19.90 & 51.61 & 63.65 & 68.81 & 59.21 & \underline{67.02} & \textbf{15.01} & 16.59 & \underline{13.70} & \textbf{16.29} \\
\end{tabular}
\end{adjustbox}
\vspace{2pt}
\begin{adjustbox}{width=0.75\linewidth}
\begin{tabular}{lcccccccccccc}
\midrule
 & \multicolumn{4}{c}{\textit{BWC Weapon Out}} & \multicolumn{4}{c}{\textit{Any Officer Handcuffing}} & \multicolumn{4}{c}{\textit{Other Officer Medical Treatment}} \\
\cmidrule(lr){2-5} \cmidrule(lr){6-9} \cmidrule(lr){10-13}
Model & F1 & P & R & AP & F1 & P & R & AP & F1 & P & R & AP \\
\midrule
\textit{Random Baseline} & 12.15 & 6.47 & \textbf{100.00} & 6.47 & 0.62 & 0.31 & \textbf{100.00} & 0.31 & 6.94 & 3.60 & \textbf{100.00} & 3.60 \\
CLIP \cite{clip} & 30.72 & \textbf{71.53} & 19.56 & 43.58 & 4.06 & \underline{62.16} & 2.10 & 12.16 & \textbf{36.39} & \underline{69.41} & \underline{24.66} & \textbf{49.88} \\
DINOv2 \cite{dinov2} & \textbf{42.94} & 65.73 & \underline{31.89} & \textbf{47.67} & \underline{13.90} & 56.13 & 7.93 & \underline{14.13} & \underline{31.37} & 68.58 & 20.33 & 44.67 \\
Hiera \cite{hiera} & 29.96 & 60.55 & 19.91 & 34.85 & 6.87 & 24.04 & 4.01 & 6.87 & 28.51 & 66.42 & 18.15 & 35.34 \\
X-CLIP \cite{xclip} & 26.26 & 59.68 & 16.83 & 35.80 & 9.95 & \textbf{66.29} & 5.38 & \textbf{15.98} & 28.66 & \textbf{72.04} & 17.89 & \underline{47.00} \\
VideoMAE v2 \cite{videomaev2} & \underline{40.02} & \underline{65.85} & 28.75 & \underline{44.65} & \textbf{15.76} & 41.00 & \underline{9.75} & 13.35 & 25.05 & 64.80 & 15.52 & 39.80 \\
\end{tabular}
\end{adjustbox}
\vspace{2pt}
\begin{adjustbox}{width=0.75\linewidth}
\begin{tabular}{lcccccccccccc}
\midrule
 & \multicolumn{4}{c}{\textit{Other Officer Physical Interaction}} & \multicolumn{4}{c}{\textit{Civilian Injured}} & \multicolumn{4}{c}{\textit{Civilian On Ground}} \\
\cmidrule(lr){2-5} \cmidrule(lr){6-9} \cmidrule(lr){10-13}
Model & F1 & P & R & AP & F1 & P & R & AP & F1 & P & R & AP \\
\midrule
\textit{Random Baseline} & 12.33 & 6.57 & \textbf{100.00} & 6.57 & 11.24 & 5.96 & \textbf{100.00} & 5.96 & 18.23 & 10.03 & \textbf{100.00} & 10.03 \\
CLIP \cite{clip} & \textbf{52.64} & \underline{74.59} & \underline{40.67} & \underline{59.47} & \textbf{56.33} & \underline{81.11} & \underline{43.15} & \textbf{67.16} & \textbf{66.04} & 78.36 & \underline{57.07} & \underline{70.96} \\
DINOv2 \cite{dinov2} & \underline{51.00} & 72.32 & 39.39 & 56.58 & \underline{46.54} & 78.55 & 33.07 & 60.39 & \underline{65.60} & 78.21 & 56.49 & \textbf{71.11} \\
Hiera \cite{hiera} & 42.93 & 65.80 & 31.86 & 46.92 & 38.05 & 66.75 & 26.61 & 46.24 & 55.01 & 65.76 & 47.28 & 58.46 \\
X-CLIP \cite{xclip} & 50.97 & \textbf{75.40} & 38.50 & \textbf{59.84} & 45.16 & \textbf{82.11} & 31.15 & \underline{62.74} & 64.39 & \underline{79.15} & 54.27 & 69.36 \\
VideoMAE v2 \cite{videomaev2} & 49.68 & 73.41 & 37.54 & 58.73 & 39.68 & 76.84 & 26.74 & 57.50 & 61.24 & \textbf{80.83} & 49.29 & 69.20 \\
\bottomrule
\end{tabular}
\end{adjustbox}
\end{center}
\vspace{6pt}
\subsection{Per-Class Results: ID Split, 10s Clips}
\label{supp:per_class_id_10s}

\begin{center}
\captionof{table}{Per-class F1, Precision (P), Recall (R), AP: all models (ID, 10s).}
\label{tab:per_class_id_10s}

\begin{adjustbox}{width=0.75\linewidth}
\begin{tabular}{lcccccccccccc}
\toprule
 & \multicolumn{4}{c}{\textit{BWC Medical Treatment}} & \multicolumn{4}{c}{\textit{BWC Physical Interaction}} & \multicolumn{4}{c}{\textit{BWC Running}} \\
\cmidrule(lr){2-5} \cmidrule(lr){6-9} \cmidrule(lr){10-13}
Model & F1 & P & R & AP & F1 & P & R & AP & F1 & P & R & AP \\
\midrule
\textit{Random Baseline} & 3.78 & 1.92 & \textbf{100.00} & 1.92 & 8.39 & 4.38 & \textbf{100.00} & 4.38 & 2.08 & 1.05 & \textbf{100.00} & 1.05 \\
CLIP \cite{clip} & 53.16 & 59.96 & 47.74 & \underline{61.60} & 68.88 & \textbf{67.51} & 70.31 & 71.89 & 8.33 & \textbf{32.14} & 4.79 & 14.60 \\
DINOv2 \cite{dinov2} & \textbf{56.04} & \textbf{68.12} & 47.60 & \textbf{65.06} & \textbf{70.78} & 63.29 & \underline{80.29} & \textbf{76.11} & 5.37 & 16.90 & 3.19 & 8.97 \\
Hiera \cite{hiera} & \underline{55.05} & 58.52 & \underline{51.97} & 56.83 & 66.96 & 58.66 & 77.99 & 73.72 & 16.06 & 23.35 & 12.23 & 12.14 \\
X-CLIP \cite{xclip} & 51.12 & \underline{66.59} & 41.48 & 59.67 & \underline{70.08} & \underline{65.88} & 74.86 & 73.39 & \underline{16.83} & \underline{29.93} & 11.70 & \underline{15.20} \\
VideoMAE v2 \cite{videomaev2} & 49.60 & 64.79 & 40.17 & 57.18 & 67.63 & 58.97 & 79.27 & \underline{74.15} & \textbf{29.72} & 27.42 & \underline{32.45} & \textbf{23.54} \\
\end{tabular}
\end{adjustbox}
\vspace{2pt}
\begin{adjustbox}{width=0.75\linewidth}
\begin{tabular}{lcccccccccccc}
\midrule
 & \multicolumn{4}{c}{\textit{BWC Weapon Out}} & \multicolumn{4}{c}{\textit{Any Officer Handcuffing}} & \multicolumn{4}{c}{\textit{Other Officer Medical Treatment}} \\
\cmidrule(lr){2-5} \cmidrule(lr){6-9} \cmidrule(lr){10-13}
Model & F1 & P & R & AP & F1 & P & R & AP & F1 & P & R & AP \\
\midrule
\textit{Random Baseline} & 17.70 & 9.71 & \textbf{100.00} & 9.71 & 1.31 & 0.66 & \textbf{100.00} & 0.66 & 10.39 & 5.48 & \textbf{100.00} & 5.48 \\
CLIP \cite{clip} & 44.18 & \textbf{67.02} & 32.95 & 52.12 & 10.12 & \underline{61.90} & 5.51 & 15.52 & \textbf{50.36} & 69.71 & \underline{39.42} & \textbf{60.53} \\
DINOv2 \cite{dinov2} & \underline{52.59} & \underline{61.35} & 46.02 & \textbf{56.77} & \underline{22.30} & 55.00 & 13.98 & 19.66 & 46.82 & \underline{72.33} & 34.61 & 55.67 \\
Hiera \cite{hiera} & 44.28 & 51.61 & 38.78 & 46.32 & 12.94 & 27.40 & 8.47 & 11.75 & 42.90 & 61.57 & 32.92 & 45.14 \\
X-CLIP \cite{xclip} & 43.13 & 60.82 & 33.41 & 47.06 & 18.71 & \textbf{61.90} & 11.02 & \underline{20.01} & \underline{49.60} & \textbf{72.35} & 37.73 & \underline{57.90} \\
VideoMAE v2 \cite{videomaev2} & \textbf{54.54} & 58.67 & \underline{50.95} & \underline{55.69} & \textbf{25.65} & 33.56 & \underline{20.76} & \textbf{20.35} & 45.74 & 67.03 & 34.71 & 53.11 \\
\end{tabular}
\end{adjustbox}
\vspace{2pt}
\begin{adjustbox}{width=0.75\linewidth}
\begin{tabular}{lcccccccccccc}
\midrule
 & \multicolumn{4}{c}{\textit{Other Officer Physical Interaction}} & \multicolumn{4}{c}{\textit{Civilian Injured}} & \multicolumn{4}{c}{\textit{Civilian On Ground}} \\
\cmidrule(lr){2-5} \cmidrule(lr){6-9} \cmidrule(lr){10-13}
Model & F1 & P & R & AP & F1 & P & R & AP & F1 & P & R & AP \\
\midrule
\textit{Random Baseline} & 18.50 & 10.19 & \textbf{100.00} & 10.19 & 15.37 & 8.33 & \textbf{100.00} & 8.33 & 23.96 & 13.61 & \textbf{100.00} & 13.61 \\
CLIP \cite{clip} & \underline{65.04} & \textbf{77.17} & 56.21 & 71.27 & \textbf{65.11} & \underline{76.14} & \underline{56.88} & \textbf{72.61} & \textbf{71.69} & \textbf{71.86} & \underline{71.52} & \underline{77.30} \\
DINOv2 \cite{dinov2} & 63.55 & 74.82 & 55.22 & 68.91 & 57.00 & 73.07 & 46.72 & 65.44 & 69.75 & 69.91 & 69.58 & \textbf{77.34} \\
Hiera \cite{hiera} & 56.19 & 62.04 & 51.35 & 60.45 & 50.96 & 61.47 & 43.53 & 53.44 & 57.83 & 52.57 & 64.25 & 65.37 \\
X-CLIP \cite{xclip} & 64.57 & \underline{75.59} & 56.35 & \textbf{71.85} & 59.15 & \textbf{78.02} & 47.63 & \underline{69.84} & \underline{70.52} & \underline{70.65} & 70.38 & 76.15 \\
VideoMAE v2 \cite{videomaev2} & \textbf{65.18} & 73.42 & \underline{58.60} & \underline{71.68} & \underline{59.52} & 74.06 & 49.75 & 67.69 & 68.54 & 70.04 & 67.11 & 75.72 \\
\bottomrule
\end{tabular}
\end{adjustbox}
\end{center}
\newpage
\subsection{Per-Class Results: ID Split, 1min Clips}
\label{supp:per_class_id_1min}

\begin{center}
\captionof{table}{Per-class F1, Precision (P), Recall (R), AP: all models (ID, 1min).}
\label{tab:per_class_id_1min}

\begin{adjustbox}{width=0.75\linewidth}
\begin{tabular}{lcccccccccccc}
\toprule
 & \multicolumn{4}{c}{\textit{BWC Medical Treatment}} & \multicolumn{4}{c}{\textit{BWC Physical Interaction}} & \multicolumn{4}{c}{\textit{BWC Running}} \\
\cmidrule(lr){2-5} \cmidrule(lr){6-9} \cmidrule(lr){10-13}
Model & F1 & P & R & AP & F1 & P & R & AP & F1 & P & R & AP \\
\midrule
\textit{Random Baseline} & 5.77 & 2.97 & \textbf{100.00} & 2.97 & 14.36 & 7.73 & \textbf{100.00} & 7.73 & 7.02 & 3.64 & \textbf{100.00} & 3.64 \\
CLIP \cite{clip} & 62.53 & 59.31 & 66.12 & 66.27 & \textbf{69.94} & \textbf{64.59} & 76.26 & 75.44 & 12.88 & \textbf{42.50} & 7.59 & 23.51 \\
DINOv2 \cite{dinov2} & \textbf{64.25} & \underline{65.71} & 62.84 & \textbf{70.65} & 68.89 & 57.12 & 86.76 & \underline{79.57} & 8.70 & 23.08 & 5.36 & 16.17 \\
Hiera \cite{hiera} & \underline{63.13} & 54.58 & \underline{74.86} & 64.82 & 62.36 & 48.87 & 86.13 & 76.91 & 18.91 & 26.40 & 14.73 & 19.72 \\
X-CLIP \cite{xclip} & 62.43 & \textbf{66.26} & 59.02 & \underline{66.69} & \underline{69.26} & \underline{59.76} & 82.35 & 77.58 & \underline{23.46} & \underline{38.00} & 16.96 & \underline{25.93} \\
VideoMAE v2 \cite{videomaev2} & 62.47 & 62.64 & 62.30 & 66.44 & 64.98 & 51.08 & \underline{89.29} & \textbf{80.21} & \textbf{39.35} & 36.06 & \underline{43.30} & \textbf{31.93} \\
\end{tabular}
\end{adjustbox}
\vspace{2pt}
\begin{adjustbox}{width=0.75\linewidth}
\begin{tabular}{lcccccccccccc}
\midrule
 & \multicolumn{4}{c}{\textit{BWC Weapon Out}} & \multicolumn{4}{c}{\textit{Any Officer Handcuffing}} & \multicolumn{4}{c}{\textit{Other Officer Medical Treatment}} \\
\cmidrule(lr){2-5} \cmidrule(lr){6-9} \cmidrule(lr){10-13}
Model & F1 & P & R & AP & F1 & P & R & AP & F1 & P & R & AP \\
\midrule
\textit{Random Baseline} & 28.69 & 16.75 & \textbf{100.00} & 16.75 & 4.17 & 2.13 & \textbf{100.00} & 2.13 & 16.68 & 9.10 & \textbf{100.00} & 9.10 \\
CLIP \cite{clip} & 53.45 & \textbf{63.47} & 46.17 & 60.27 & 11.03 & \textbf{57.14} & 6.11 & 19.95 & \textbf{61.25} & \textbf{74.94} & 51.79 & \textbf{70.47} \\
DINOv2 \cite{dinov2} & \underline{57.56} & 56.31 & 58.87 & \underline{66.17} & \underline{25.73} & 55.00 & 16.79 & 24.82 & 57.74 & \underline{74.93} & 46.96 & 65.52 \\
Hiera \cite{hiera} & 52.44 & 46.67 & 59.84 & 56.44 & 14.86 & 29.55 & 9.92 & 18.26 & 53.41 & 59.87 & 48.21 & 56.83 \\
X-CLIP \cite{xclip} & 54.63 & \underline{59.72} & 50.34 & 58.24 & 20.00 & \underline{55.17} & 12.21 & \underline{26.01} & 60.63 & 73.85 & 51.43 & \underline{66.47} \\
VideoMAE v2 \cite{videomaev2} & \textbf{59.54} & 52.44 & \underline{68.87} & \textbf{66.74} & \textbf{30.96} & 34.26 & \underline{28.24} & \textbf{28.91} & \underline{60.77} & 72.82 & \underline{52.14} & 64.44 \\
\end{tabular}
\end{adjustbox}
\vspace{2pt}
\begin{adjustbox}{width=0.75\linewidth}
\begin{tabular}{lcccccccccccc}
\midrule
 & \multicolumn{4}{c}{\textit{Other Officer Physical Interaction}} & \multicolumn{4}{c}{\textit{Civilian Injured}} & \multicolumn{4}{c}{\textit{Civilian On Ground}} \\
\cmidrule(lr){2-5} \cmidrule(lr){6-9} \cmidrule(lr){10-13}
Model & F1 & P & R & AP & F1 & P & R & AP & F1 & P & R & AP \\
\midrule
\textit{Random Baseline} & 29.78 & 17.50 & \textbf{100.00} & 17.50 & 22.95 & 12.97 & \textbf{100.00} & 12.97 & 33.78 & 20.32 & \textbf{100.00} & 20.32 \\
CLIP \cite{clip} & 70.04 & \textbf{76.83} & 64.35 & 79.45 & \textbf{69.92} & \underline{72.60} & \underline{67.42} & \textbf{76.38} & \textbf{70.75} & \textbf{63.76} & 79.46 & \textbf{81.91} \\
DINOv2 \cite{dinov2} & 70.14 & \underline{76.03} & 65.09 & 78.31 & 64.82 & 72.27 & 58.77 & 69.94 & 69.88 & 62.47 & 79.30 & \underline{81.85} \\
Hiera \cite{hiera} & 62.19 & 58.28 & 66.67 & 70.73 & 55.16 & 54.46 & 55.89 & 57.27 & 56.62 & 44.27 & 78.50 & 71.12 \\
X-CLIP \cite{xclip} & \underline{70.30} & 75.43 & 65.83 & \underline{79.88} & 67.60 & \textbf{76.75} & 60.40 & 74.36 & \underline{70.33} & \underline{62.54} & \underline{80.34} & 81.62 \\
VideoMAE v2 \cite{videomaev2} & \textbf{70.92} & 72.14 & \underline{69.73} & \textbf{80.38} & \underline{68.84} & 71.90 & 66.04 & \underline{74.70} & 67.51 & 60.01 & 77.14 & 80.04 \\
\bottomrule
\end{tabular}
\end{adjustbox}
\end{center}
\vspace{6pt}
\subsection{Per-Class Results: OOD-T Split, 1s Clips}
\label{supp:per_class_ood_t_1s}

\begin{center}
\captionof{table}{Per-class F1, Precision (P), Recall (R), AP: all models (OOD-T, 1s).}
\label{tab:per_class_ood_t_1s}

\begin{adjustbox}{width=0.75\linewidth}
\begin{tabular}{lcccccccccccc}
\toprule
 & \multicolumn{4}{c}{\textit{BWC Medical Treatment}} & \multicolumn{4}{c}{\textit{BWC Physical Interaction}} & \multicolumn{4}{c}{\textit{BWC Running}} \\
\cmidrule(lr){2-5} \cmidrule(lr){6-9} \cmidrule(lr){10-13}
Model & F1 & P & R & AP & F1 & P & R & AP & F1 & P & R & AP \\
\midrule
\textit{Random Baseline} & 5.86 & 3.02 & \textbf{100.00} & 3.02 & 14.34 & 7.73 & \textbf{100.00} & 7.73 & 6.68 & 3.46 & \textbf{100.00} & 3.46 \\
CLIP \cite{clip} & \textbf{65.40} & 65.82 & \underline{64.97} & \underline{70.77} & 57.80 & \textbf{87.75} & 43.09 & \underline{61.53} & 1.19 & \underline{100.00} & 0.60 & \underline{54.71} \\
DINOv2 \cite{dinov2} & \underline{63.63} & \underline{68.31} & 59.55 & 64.33 & \underline{58.57} & 85.38 & 44.57 & 59.66 & 1.93 & \textbf{100.00} & 0.98 & 43.89 \\
Hiera \cite{hiera} & 60.23 & \textbf{74.74} & 50.43 & \textbf{72.86} & 56.32 & 81.77 & 42.96 & 57.79 & \underline{10.67} & 81.72 & 5.71 & 54.49 \\
X-CLIP \cite{xclip} & 58.22 & 67.46 & 51.20 & 66.62 & \textbf{58.82} & \underline{86.08} & \underline{44.67} & \textbf{64.40} & 4.11 & 93.33 & 2.10 & 51.96 \\
VideoMAE v2 \cite{videomaev2} & 56.00 & 67.68 & 47.76 & 68.64 & 56.18 & 82.89 & 42.49 & 59.82 & \textbf{48.88} & 96.88 & \underline{32.68} & \textbf{85.36} \\
\end{tabular}
\end{adjustbox}
\vspace{2pt}
\begin{adjustbox}{width=0.75\linewidth}
\begin{tabular}{lcccccccccccc}
\midrule
 & \multicolumn{4}{c}{\textit{BWC Weapon Out}} & \multicolumn{4}{c}{\textit{Any Officer Handcuffing}} & \multicolumn{4}{c}{\textit{Other Officer Medical Treatment}} \\
\cmidrule(lr){2-5} \cmidrule(lr){6-9} \cmidrule(lr){10-13}
Model & F1 & P & R & AP & F1 & P & R & AP & F1 & P & R & AP \\
\midrule
\textit{Random Baseline} & 5.11 & 2.62 & \textbf{100.00} & 2.62 & 0.21 & 0.11 & \textbf{100.00} & 0.11 & 14.67 & 7.92 & \textbf{100.00} & 7.92 \\
CLIP \cite{clip} & 33.01 & 48.37 & 25.05 & 32.92 & 0.00 & 0.00 & 0.00 & 1.05 & 42.04 & \underline{72.57} & 29.59 & \underline{59.09} \\
DINOv2 \cite{dinov2} & \textbf{49.15} & \textbf{60.70} & 41.29 & \textbf{50.11} & \underline{3.51} & \underline{6.25} & 2.44 & \underline{1.62} & \textbf{46.59} & 72.23 & \underline{34.38} & 56.81 \\
Hiera \cite{hiera} & 35.52 & 29.04 & \underline{45.74} & 38.83 & 0.00 & 0.00 & 0.00 & 0.62 & \underline{44.27} & 69.18 & 32.55 & 50.84 \\
X-CLIP \cite{xclip} & 36.88 & 47.87 & 30.00 & 35.53 & 0.00 & 0.00 & 0.00 & 1.02 & 37.40 & \textbf{80.65} & 24.34 & \textbf{60.29} \\
VideoMAE v2 \cite{videomaev2} & \underline{48.07} & \underline{51.65} & 44.95 & \underline{47.05} & \textbf{10.53} & \textbf{18.75} & \underline{7.32} & \textbf{5.31} & 43.75 & 69.81 & 31.86 & 56.67 \\
\end{tabular}
\end{adjustbox}
\vspace{2pt}
\begin{adjustbox}{width=0.75\linewidth}
\begin{tabular}{lcccccccccccc}
\midrule
 & \multicolumn{4}{c}{\textit{Other Officer Physical Interaction}} & \multicolumn{4}{c}{\textit{Civilian Injured}} & \multicolumn{4}{c}{\textit{Civilian On Ground}} \\
\cmidrule(lr){2-5} \cmidrule(lr){6-9} \cmidrule(lr){10-13}
Model & F1 & P & R & AP & F1 & P & R & AP & F1 & P & R & AP \\
\midrule
\textit{Random Baseline} & 26.42 & 15.22 & \textbf{100.00} & 15.22 & 19.43 & 10.76 & \textbf{100.00} & 10.76 & 32.63 & 19.50 & \textbf{100.00} & 19.50 \\
CLIP \cite{clip} & \textbf{52.92} & \underline{85.71} & 38.28 & \textbf{70.83} & \textbf{61.92} & \underline{87.81} & 47.82 & \textbf{73.35} & \textbf{64.49} & 88.50 & \underline{50.73} & \textbf{78.02} \\
DINOv2 \cite{dinov2} & 50.00 & 81.96 & 35.97 & 65.16 & 58.61 & 85.25 & 44.65 & 69.74 & \underline{63.10} & \underline{89.24} & 48.81 & 74.96 \\
Hiera \cite{hiera} & 49.82 & 68.02 & \underline{39.30} & 55.82 & 56.48 & 69.40 & 47.62 & 62.32 & 54.40 & 64.78 & 46.89 & 63.40 \\
X-CLIP \cite{xclip} & 48.06 & \textbf{88.01} & 33.05 & \underline{68.73} & 52.83 & \textbf{87.82} & 37.77 & 69.06 & 62.40 & \textbf{90.46} & 47.62 & \underline{77.02} \\
VideoMAE v2 \cite{videomaev2} & \underline{50.41} & 83.81 & 36.04 & 68.19 & \underline{60.78} & 81.49 & \underline{48.47} & \underline{70.41} & 61.92 & 87.04 & 48.05 & 76.39 \\
\bottomrule
\end{tabular}
\end{adjustbox}
\end{center}
\newpage
\subsection{Per-Class Results: OOD-T Split, 10s Clips}
\label{supp:per_class_ood_t_10s}

\begin{center}
\captionof{table}{Per-class F1, Precision (P), Recall (R), AP: all models (OOD-T, 10s).}
\label{tab:per_class_ood_t_10s}

\begin{adjustbox}{width=0.75\linewidth}
\begin{tabular}{lcccccccccccc}
\toprule
 & \multicolumn{4}{c}{\textit{BWC Medical Treatment}} & \multicolumn{4}{c}{\textit{BWC Physical Interaction}} & \multicolumn{4}{c}{\textit{BWC Running}} \\
\cmidrule(lr){2-5} \cmidrule(lr){6-9} \cmidrule(lr){10-13}
Model & F1 & P & R & AP & F1 & P & R & AP & F1 & P & R & AP \\
\midrule
\textit{Random Baseline} & 8.63 & 4.51 & \textbf{100.00} & 4.51 & 22.72 & 12.82 & \textbf{100.00} & 12.82 & 11.14 & 5.90 & \textbf{100.00} & 5.90 \\
CLIP \cite{clip} & \underline{75.77} & 74.32 & 77.27 & \underline{81.54} & \textbf{60.35} & \textbf{90.76} & 45.20 & \underline{63.81} & 3.42 & \underline{100.00} & 1.74 & \underline{63.94} \\
DINOv2 \cite{dinov2} & 74.12 & \underline{76.83} & 71.59 & 76.23 & 59.46 & 83.39 & 46.20 & 63.10 & 5.08 & \textbf{100.00} & 2.61 & 55.00 \\
Hiera \cite{hiera} & \textbf{81.59} & \textbf{81.36} & \underline{81.82} & \textbf{86.46} & 55.20 & 72.40 & 44.60 & 60.06 & \underline{30.66} & 77.19 & 19.13 & 60.85 \\
X-CLIP \cite{xclip} & 68.73 & 75.51 & 63.07 & 79.88 & \underline{60.21} & \underline{87.12} & 46.00 & \textbf{66.33} & 12.20 & 93.75 & 6.52 & 63.18 \\
VideoMAE v2 \cite{videomaev2} & 71.30 & 72.78 & 69.89 & 79.31 & 58.51 & 75.39 & \underline{47.80} & 61.91 & \textbf{72.28} & 96.38 & \underline{57.83} & \textbf{89.58} \\
\end{tabular}
\end{adjustbox}
\vspace{2pt}
\begin{adjustbox}{width=0.75\linewidth}
\begin{tabular}{lcccccccccccc}
\midrule
 & \multicolumn{4}{c}{\textit{BWC Weapon Out}} & \multicolumn{4}{c}{\textit{Any Officer Handcuffing}} & \multicolumn{4}{c}{\textit{Other Officer Medical Treatment}} \\
\cmidrule(lr){2-5} \cmidrule(lr){6-9} \cmidrule(lr){10-13}
Model & F1 & P & R & AP & F1 & P & R & AP & F1 & P & R & AP \\
\midrule
\textit{Random Baseline} & 8.68 & 4.54 & \textbf{100.00} & 4.54 & 0.61 & 0.31 & \textbf{100.00} & 0.31 & 23.92 & 13.59 & \textbf{100.00} & 13.59 \\
CLIP \cite{clip} & 41.61 & 35.77 & 49.72 & 38.03 & 0.00 & 0.00 & 0.00 & 4.61 & \underline{57.54} & 75.69 & 46.42 & 69.07 \\
DINOv2 \cite{dinov2} & \textbf{49.22} & \textbf{40.74} & 62.15 & \underline{55.76} & \underline{10.00} & \underline{12.50} & 8.33 & \underline{6.78} & 57.11 & \underline{75.70} & 45.85 & 65.92 \\
Hiera \cite{hiera} & 29.15 & 18.48 & 68.93 & 41.74 & 8.70 & 5.88 & 16.67 & 3.27 & 56.20 & 67.19 & 48.30 & 61.81 \\
X-CLIP \cite{xclip} & 44.10 & \underline{36.40} & 55.93 & 45.07 & 0.00 & 0.00 & 0.00 & 5.38 & 51.53 & \textbf{79.53} & 38.11 & \textbf{70.91} \\
VideoMAE v2 \cite{videomaev2} & \underline{49.08} & 36.31 & \underline{75.71} & \textbf{60.35} & \textbf{18.18} & \textbf{20.00} & \underline{16.67} & \textbf{11.57} & \textbf{60.68} & 72.32 & \underline{52.26} & \underline{69.09} \\
\end{tabular}
\end{adjustbox}
\vspace{2pt}
\begin{adjustbox}{width=0.75\linewidth}
\begin{tabular}{lcccccccccccc}
\midrule
 & \multicolumn{4}{c}{\textit{Other Officer Physical Interaction}} & \multicolumn{4}{c}{\textit{Civilian Injured}} & \multicolumn{4}{c}{\textit{Civilian On Ground}} \\
\cmidrule(lr){2-5} \cmidrule(lr){6-9} \cmidrule(lr){10-13}
Model & F1 & P & R & AP & F1 & P & R & AP & F1 & P & R & AP \\
\midrule
\textit{Random Baseline} & 34.77 & 21.05 & \textbf{100.00} & 21.05 & 27.66 & 16.05 & \textbf{100.00} & 16.05 & 43.49 & 27.79 & \textbf{100.00} & 27.79 \\
CLIP \cite{clip} & \textbf{69.01} & \underline{83.68} & 58.71 & \textbf{80.35} & \textbf{67.79} & \textbf{82.23} & 57.67 & \textbf{77.99} & \textbf{72.51} & 81.64 & 65.22 & \textbf{84.36} \\
DINOv2 \cite{dinov2} & 63.92 & 76.66 & 54.81 & 74.15 & 64.83 & 79.72 & 54.63 & 74.64 & 71.03 & \underline{82.98} & 62.08 & 82.28 \\
Hiera \cite{hiera} & 58.80 & 58.41 & \underline{59.20} & 66.16 & 60.37 & 61.01 & 59.74 & 66.90 & 59.06 & 53.52 & \underline{65.87} & 71.02 \\
X-CLIP \cite{xclip} & \underline{66.52} & \textbf{85.17} & 54.57 & \underline{78.76} & 60.30 & \underline{79.95} & 48.40 & 73.82 & \underline{71.34} & \textbf{84.00} & 61.99 & \underline{83.76} \\
VideoMAE v2 \cite{videomaev2} & 65.97 & 76.95 & 57.73 & 77.74 & \underline{66.90} & 74.75 & \underline{60.54} & \underline{74.72} & 70.50 & 79.58 & 63.28 & 82.39 \\
\bottomrule
\end{tabular}
\end{adjustbox}
\end{center}
\vspace{6pt}
\subsection{Per-Class Results: OOD-T Split, 1min Clips}
\label{supp:per_class_ood_t_1min}

\begin{center}
\captionof{table}{Per-class F1, Precision (P), Recall (R), AP: all models (OOD-T, 1min).}
\label{tab:per_class_ood_t_1min}

\begin{adjustbox}{width=0.75\linewidth}
\begin{tabular}{lcccccccccccc}
\toprule
 & \multicolumn{4}{c}{\textit{BWC Medical Treatment}} & \multicolumn{4}{c}{\textit{BWC Physical Interaction}} & \multicolumn{4}{c}{\textit{BWC Running}} \\
\cmidrule(lr){2-5} \cmidrule(lr){6-9} \cmidrule(lr){10-13}
Model & F1 & P & R & AP & F1 & P & R & AP & F1 & P & R & AP \\
\midrule
\textit{Random Baseline} & 14.85 & 8.02 & \textbf{100.00} & 8.02 & 36.92 & 22.64 & \textbf{100.00} & 22.64 & 27.23 & 15.76 & \textbf{100.00} & 15.76 \\
CLIP \cite{clip} & \underline{87.04} & \underline{90.38} & 83.93 & \textbf{94.03} & \underline{62.45} & \textbf{93.67} & 46.84 & \underline{72.28} & 5.31 & 100.00 & 2.73 & \underline{73.52} \\
DINOv2 \cite{dinov2} & 83.02 & 88.00 & 78.57 & 88.91 & 62.31 & 79.41 & 51.27 & 70.38 & 8.70 & \underline{100.00} & 4.55 & 64.16 \\
Hiera \cite{hiera} & \textbf{90.76} & 85.71 & \underline{96.43} & \underline{94.03} & 59.36 & 67.20 & 53.16 & 68.45 & \underline{45.45} & 79.55 & 31.82 & 70.07 \\
X-CLIP \cite{xclip} & 82.00 & \textbf{93.18} & 73.21 & 92.92 & \textbf{63.71} & \underline{87.78} & 50.00 & \textbf{74.01} & 21.14 & \textbf{100.00} & 11.82 & 73.38 \\
VideoMAE v2 \cite{videomaev2} & 78.18 & 79.63 & 76.79 & 88.35 & 60.50 & 69.11 & \underline{53.80} & 69.06 & \textbf{81.87} & 95.18 & \underline{71.82} & \textbf{92.10} \\
\end{tabular}
\end{adjustbox}
\vspace{2pt}
\begin{adjustbox}{width=0.75\linewidth}
\begin{tabular}{lcccccccccccc}
\midrule
 & \multicolumn{4}{c}{\textit{BWC Weapon Out}} & \multicolumn{4}{c}{\textit{Any Officer Handcuffing}} & \multicolumn{4}{c}{\textit{Other Officer Medical Treatment}} \\
\cmidrule(lr){2-5} \cmidrule(lr){6-9} \cmidrule(lr){10-13}
Model & F1 & P & R & AP & F1 & P & R & AP & F1 & P & R & AP \\
\midrule
\textit{Random Baseline} & 16.80 & 9.17 & \textbf{100.00} & 9.17 & 3.10 & 1.58 & \textbf{100.00} & 1.58 & 38.99 & 24.21 & \textbf{100.00} & 24.21 \\
CLIP \cite{clip} & \textbf{49.25} & \textbf{36.30} & 76.56 & 49.35 & 0.00 & 0.00 & 0.00 & 18.46 & \underline{67.36} & \underline{81.51} & 57.40 & \underline{79.06} \\
DINOv2 \cite{dinov2} & \underline{47.22} & \underline{33.55} & 79.69 & \underline{60.81} & \textbf{23.53} & \textbf{33.33} & 18.18 & \underline{20.56} & 64.38 & 76.42 & 55.62 & 76.59 \\
Hiera \cite{hiera} & 30.89 & 18.69 & 89.06 & 50.13 & 13.79 & 11.11 & 18.18 & 12.86 & 66.24 & 71.72 & 61.54 & 75.29 \\
X-CLIP \cite{xclip} & 46.08 & 32.68 & 78.12 & 57.13 & 0.00 & 0.00 & 0.00 & 19.13 & 63.20 & \textbf{85.00} & 50.30 & \textbf{80.28} \\
VideoMAE v2 \cite{videomaev2} & 44.62 & 29.59 & \underline{90.62} & \textbf{66.56} & \underline{20.00} & \underline{22.22} & \underline{18.18} & \textbf{22.23} & \textbf{68.17} & 74.65 & \underline{62.72} & 78.20 \\
\end{tabular}
\end{adjustbox}
\vspace{2pt}
\begin{adjustbox}{width=0.75\linewidth}
\begin{tabular}{lcccccccccccc}
\midrule
 & \multicolumn{4}{c}{\textit{Other Officer Physical Interaction}} & \multicolumn{4}{c}{\textit{Civilian Injured}} & \multicolumn{4}{c}{\textit{Civilian On Ground}} \\
\cmidrule(lr){2-5} \cmidrule(lr){6-9} \cmidrule(lr){10-13}
Model & F1 & P & R & AP & F1 & P & R & AP & F1 & P & R & AP \\
\midrule
\textit{Random Baseline} & 50.38 & 33.67 & \textbf{100.00} & 33.67 & 43.15 & 27.51 & \textbf{100.00} & 27.51 & 58.99 & 41.83 & \textbf{100.00} & 41.83 \\
CLIP \cite{clip} & \textbf{74.65} & \textbf{81.41} & 68.94 & \textbf{86.44} & \underline{73.84} & \underline{83.55} & 66.15 & \textbf{83.84} & 74.39 & 75.17 & 73.63 & \underline{87.68} \\
DINOv2 \cite{dinov2} & 71.04 & 75.85 & 66.81 & 81.68 & \textbf{75.00} & \textbf{84.87} & 67.19 & \underline{83.45} & \textbf{77.19} & \underline{79.14} & 75.34 & \textbf{88.09} \\
Hiera \cite{hiera} & 65.06 & 57.76 & \underline{74.47} & 74.66 & 67.15 & 63.01 & \underline{71.87} & 76.97 & 69.17 & 58.18 & \underline{85.27} & 79.49 \\
X-CLIP \cite{xclip} & 71.43 & \underline{81.08} & 63.83 & \underline{84.18} & 68.10 & 82.84 & 57.81 & 80.16 & \underline{76.38} & \textbf{79.34} & 73.63 & 87.43 \\
VideoMAE v2 \cite{videomaev2} & \underline{72.45} & 73.89 & 71.06 & 83.68 & 71.74 & 75.00 & 68.75 & 81.63 & 75.97 & 74.59 & 77.40 & 86.14 \\
\bottomrule
\end{tabular}
\end{adjustbox}
\end{center}
\newpage
\subsection{Per-Class Results: OOD-L Split, 1s Clips}
\label{supp:per_class_ood_l_1s}

\begin{center}
\captionof{table}{Per-class F1, Precision (P), Recall (R), AP: all models (OOD-L, 1s).}
\label{tab:per_class_ood_l_1s}

\begin{adjustbox}{width=0.75\linewidth}
\begin{tabular}{lcccccccccccc}
\toprule
 & \multicolumn{4}{c}{\textit{BWC Medical Treatment}} & \multicolumn{4}{c}{\textit{BWC Physical Interaction}} & \multicolumn{4}{c}{\textit{BWC Running}} \\
\cmidrule(lr){2-5} \cmidrule(lr){6-9} \cmidrule(lr){10-13}
Model & F1 & P & R & AP & F1 & P & R & AP & F1 & P & R & AP \\
\midrule
\textit{Random Baseline} & 4.30 & 2.20 & \textbf{100.00} & 2.20 & 8.65 & 4.52 & \textbf{100.00} & 4.52 & 2.22 & 1.12 & \textbf{100.00} & 1.12 \\
CLIP \cite{clip} & \underline{35.47} & \underline{79.84} & 22.80 & \underline{65.78} & \underline{58.71} & \textbf{80.34} & 46.26 & \textbf{68.69} & 3.18 & 11.59 & 1.84 & 13.70 \\
DINOv2 \cite{dinov2} & 2.72 & 40.00 & 1.41 & 56.85 & 58.04 & 70.69 & 49.23 & 65.43 & 0.00 & 0.00 & 0.00 & 14.24 \\
Hiera \cite{hiera} & 28.04 & 58.36 & 18.45 & 38.93 & 58.27 & 56.71 & \underline{59.92} & 58.63 & 2.24 & 41.67 & 1.15 & 13.68 \\
X-CLIP \cite{xclip} & \textbf{39.53} & \textbf{85.21} & \underline{25.73} & \textbf{78.47} & \textbf{60.88} & \underline{74.87} & 51.29 & \underline{67.68} & \underline{16.56} & \underline{81.63} & 9.22 & \underline{28.27} \\
VideoMAE v2 \cite{videomaev2} & 10.48 & 73.85 & 5.64 & 57.26 & 49.85 & 62.50 & 41.45 & 54.89 & \textbf{43.06} & \textbf{94.53} & \underline{27.88} & \textbf{63.48} \\
\end{tabular}
\end{adjustbox}
\vspace{2pt}
\begin{adjustbox}{width=0.75\linewidth}
\begin{tabular}{lcccccccccccc}
\midrule
 & \multicolumn{4}{c}{\textit{BWC Weapon Out}} & \multicolumn{4}{c}{\textit{Any Officer Handcuffing}} & \multicolumn{4}{c}{\textit{Other Officer Medical Treatment}} \\
\cmidrule(lr){2-5} \cmidrule(lr){6-9} \cmidrule(lr){10-13}
Model & F1 & P & R & AP & F1 & P & R & AP & F1 & P & R & AP \\
\midrule
\textit{Random Baseline} & 5.73 & 2.95 & \textbf{100.00} & 2.95 & 0.78 & 0.39 & \textbf{100.00} & 0.39 & \underline{6.87} & 3.56 & \textbf{100.00} & 3.56 \\
CLIP \cite{clip} & 32.55 & \textbf{69.23} & 21.28 & 40.46 & 1.31 & \textbf{100.00} & 0.66 & 11.59 & 2.24 & 29.63 & 1.16 & 15.79 \\
DINOv2 \cite{dinov2} & \textbf{42.34} & \underline{59.31} & \underline{32.92} & \underline{41.56} & 13.33 & 30.23 & 8.55 & 12.06 & 3.48 & 40.32 & 1.82 & \underline{20.47} \\
Hiera \cite{hiera} & 24.50 & 23.82 & 25.22 & 21.11 & \underline{17.22} & 31.58 & 11.84 & 14.63 & \textbf{7.37} & 23.81 & \underline{4.36} & 13.22 \\
X-CLIP \cite{xclip} & 29.53 & 56.72 & 19.96 & 33.48 & 11.83 & \underline{58.82} & 6.58 & \underline{15.15} & 2.28 & \textbf{57.14} & 1.16 & \textbf{30.61} \\
VideoMAE v2 \cite{videomaev2} & \underline{38.32} & 56.18 & 29.07 & \textbf{41.67} & \textbf{19.49} & 44.19 & \underline{12.50} & \textbf{20.80} & 4.55 & \underline{44.59} & 2.40 & 16.86 \\
\end{tabular}
\end{adjustbox}
\vspace{2pt}
\begin{adjustbox}{width=0.75\linewidth}
\begin{tabular}{lcccccccccccc}
\midrule
 & \multicolumn{4}{c}{\textit{Other Officer Physical Interaction}} & \multicolumn{4}{c}{\textit{Civilian Injured}} & \multicolumn{4}{c}{\textit{Civilian On Ground}} \\
\cmidrule(lr){2-5} \cmidrule(lr){6-9} \cmidrule(lr){10-13}
Model & F1 & P & R & AP & F1 & P & R & AP & F1 & P & R & AP \\
\midrule
\textit{Random Baseline} & 12.38 & 6.60 & \textbf{100.00} & 6.60 & 5.24 & 2.69 & \textbf{100.00} & 2.69 & 13.79 & 7.40 & \textbf{100.00} & 7.40 \\
CLIP \cite{clip} & 25.08 & \textbf{72.61} & 15.16 & \textbf{49.86} & \textbf{40.75} & \underline{81.32} & \underline{27.19} & 51.95 & \underline{57.36} & \textbf{79.65} & 44.82 & \underline{62.75} \\
DINOv2 \cite{dinov2} & \textbf{32.60} & 68.51 & 21.39 & 42.05 & 22.10 & 66.35 & 13.26 & 38.93 & 56.47 & 73.63 & 45.79 & \textbf{64.95} \\
Hiera \cite{hiera} & \underline{27.59} & 32.93 & \underline{23.74} & 23.34 & 28.38 & 31.05 & 26.13 & 21.91 & 41.13 & 36.75 & \underline{46.70} & 43.19 \\
X-CLIP \cite{xclip} & 23.77 & \underline{69.58} & 14.34 & \underline{44.87} & 35.40 & \textbf{87.50} & 22.19 & \textbf{53.27} & \textbf{57.78} & \underline{79.21} & 45.48 & 62.28 \\
VideoMAE v2 \cite{videomaev2} & 26.18 & 60.68 & 16.69 & 37.97 & \underline{35.72} & 77.07 & 23.25 & \underline{52.74} & 50.68 & 71.43 & 39.27 & 51.63 \\
\bottomrule
\end{tabular}
\end{adjustbox}
\end{center}
\vspace{6pt}
\subsection{Per-Class Results: OOD-L Split, 10s Clips}
\label{supp:per_class_ood_l_10s}

\begin{center}
\captionof{table}{Per-class F1, Precision (P), Recall (R), AP: all models (OOD-L, 10s).}
\label{tab:per_class_ood_l_10s}

\begin{adjustbox}{width=0.75\linewidth}
\begin{tabular}{lcccccccccccc}
\toprule
 & \multicolumn{4}{c}{\textit{BWC Medical Treatment}} & \multicolumn{4}{c}{\textit{BWC Physical Interaction}} & \multicolumn{4}{c}{\textit{BWC Running}} \\
\cmidrule(lr){2-5} \cmidrule(lr){6-9} \cmidrule(lr){10-13}
Model & F1 & P & R & AP & F1 & P & R & AP & F1 & P & R & AP \\
\midrule
\textit{Random Baseline} & 4.50 & 2.30 & \textbf{100.00} & 2.30 & 10.88 & 5.75 & \textbf{100.00} & 5.75 & 4.40 & 2.25 & \textbf{100.00} & 2.25 \\
CLIP \cite{clip} & \underline{60.40} & \underline{76.27} & 50.00 & 67.62 & \underline{66.97} & \textbf{69.71} & 64.44 & \textbf{72.28} & 14.41 & 34.78 & 9.09 & 22.15 \\
DINOv2 \cite{dinov2} & 16.51 & 47.37 & 10.00 & 62.55 & \textbf{67.88} & 62.22 & 74.67 & \underline{70.35} & 0.00 & 0.00 & 0.00 & 19.88 \\
Hiera \cite{hiera} & 50.85 & 51.72 & 50.00 & 46.78 & 58.98 & 44.47 & \underline{87.56} & 67.63 & 10.31 & 55.56 & 5.68 & 22.81 \\
X-CLIP \cite{xclip} & \textbf{70.13} & \textbf{84.37} & \underline{60.00} & \textbf{83.63} & 66.10 & \underline{63.16} & 69.33 & 70.23 & \underline{29.09} & \underline{72.73} & 18.18 & \underline{33.44} \\
VideoMAE v2 \cite{videomaev2} & 39.37 & 67.57 & 27.78 & \underline{67.83} & 62.87 & 53.61 & 76.00 & 61.44 & \textbf{55.56} & \textbf{92.11} & \underline{39.77} & \textbf{62.40} \\
\end{tabular}
\end{adjustbox}
\vspace{2pt}
\begin{adjustbox}{width=0.75\linewidth}
\begin{tabular}{lcccccccccccc}
\midrule
 & \multicolumn{4}{c}{\textit{BWC Weapon Out}} & \multicolumn{4}{c}{\textit{Any Officer Handcuffing}} & \multicolumn{4}{c}{\textit{Other Officer Medical Treatment}} \\
\cmidrule(lr){2-5} \cmidrule(lr){6-9} \cmidrule(lr){10-13}
Model & F1 & P & R & AP & F1 & P & R & AP & F1 & P & R & AP \\
\midrule
\textit{Random Baseline} & 7.82 & 4.07 & \textbf{100.00} & 4.07 & 1.87 & 0.95 & \textbf{100.00} & 0.95 & 8.19 & 4.27 & \textbf{100.00} & 4.27 \\
CLIP \cite{clip} & \textbf{49.13} & \textbf{54.62} & 44.65 & 48.40 & 5.26 & \textbf{100.00} & 2.70 & \underline{30.59} & 6.25 & 24.00 & 3.59 & 21.61 \\
DINOv2 \cite{dinov2} & 46.24 & 42.78 & 50.31 & \textbf{51.30} & 19.23 & 33.33 & 13.51 & 23.19 & 12.50 & \underline{48.00} & 7.19 & \underline{27.07} \\
Hiera \cite{hiera} & 26.35 & 18.01 & 49.06 & 25.79 & 27.12 & 36.36 & 21.62 & 28.98 & \textbf{22.92} & 27.27 & \underline{19.76} & 18.44 \\
X-CLIP \cite{xclip} & \underline{47.59} & \underline{45.66} & 49.69 & 45.38 & \underline{29.79} & \underline{70.00} & 18.92 & 30.14 & 9.94 & \textbf{64.29} & 5.39 & \textbf{42.12} \\
VideoMAE v2 \cite{videomaev2} & 45.77 & 37.86 & \underline{57.86} & \underline{48.78} & \textbf{34.48} & 47.62 & \underline{27.03} & \textbf{35.34} & \underline{18.52} & 40.82 & 11.98 & 24.79 \\
\end{tabular}
\end{adjustbox}
\vspace{2pt}
\begin{adjustbox}{width=0.75\linewidth}
\begin{tabular}{lcccccccccccc}
\midrule
 & \multicolumn{4}{c}{\textit{Other Officer Physical Interaction}} & \multicolumn{4}{c}{\textit{Civilian Injured}} & \multicolumn{4}{c}{\textit{Civilian On Ground}} \\
\cmidrule(lr){2-5} \cmidrule(lr){6-9} \cmidrule(lr){10-13}
Model & F1 & P & R & AP & F1 & P & R & AP & F1 & P & R & AP \\
\midrule
\textit{Random Baseline} & 16.30 & 8.87 & \textbf{100.00} & 8.87 & 5.67 & 2.92 & \textbf{100.00} & 2.92 & 16.30 & 8.87 & \textbf{100.00} & 8.87 \\
CLIP \cite{clip} & 43.63 & \textbf{66.08} & 32.56 & \textbf{54.75} & 56.73 & \underline{62.77} & 51.75 & \textbf{67.86} & \textbf{62.74} & \textbf{64.16} & 61.38 & \underline{66.75} \\
DINOv2 \cite{dinov2} & \underline{44.90} & 54.77 & 38.04 & 47.00 & 50.00 & 58.14 & 43.86 & 53.93 & \underline{60.56} & 56.16 & \underline{65.71} & \textbf{69.81} \\
Hiera \cite{hiera} & 35.55 & 27.53 & \underline{50.14} & 29.02 & 36.27 & 25.17 & 64.91 & 28.38 & 31.82 & 21.31 & 62.82 & 48.21 \\
X-CLIP \cite{xclip} & 41.13 & \underline{59.56} & 31.41 & \underline{49.52} & \underline{58.95} & \textbf{73.68} & 49.12 & \underline{67.78} & 59.86 & \underline{58.15} & 61.67 & 65.71 \\
VideoMAE v2 \cite{videomaev2} & \textbf{46.91} & 53.93 & 41.50 & 46.99 & \textbf{62.76} & 60.00 & \underline{65.79} & 67.70 & 49.09 & 44.68 & 54.47 & 56.71 \\
\bottomrule
\end{tabular}
\end{adjustbox}
\end{center}
\newpage
\subsection{Per-Class Results: OOD-L Split, 1min Clips}
\label{supp:per_class_ood_l_1min}

\begin{center}
\captionof{table}{Per-class F1, Precision (P), Recall (R), AP: all models (OOD-L, 1min).}
\label{tab:per_class_ood_l_1min}

\begin{adjustbox}{width=0.75\linewidth}
\begin{tabular}{lcccccccccccc}
\toprule
 & \multicolumn{4}{c}{\textit{BWC Medical Treatment}} & \multicolumn{4}{c}{\textit{BWC Physical Interaction}} & \multicolumn{4}{c}{\textit{BWC Running}} \\
\cmidrule(lr){2-5} \cmidrule(lr){6-9} \cmidrule(lr){10-13}
Model & F1 & P & R & AP & F1 & P & R & AP & F1 & P & R & AP \\
\midrule
\textit{Random Baseline} & 5.12 & 2.63 & \textbf{100.00} & 2.63 & 15.61 & 8.47 & \textbf{100.00} & 8.47 & 13.35 & 7.15 & \textbf{100.00} & 7.15 \\
CLIP \cite{clip} & \underline{78.95} & \underline{75.00} & 83.33 & 68.26 & \textbf{66.67} & \textbf{57.50} & 79.31 & \underline{76.58} & 21.05 & 75.00 & 12.24 & 43.49 \\
DINOv2 \cite{dinov2} & 41.38 & 54.55 & 33.33 & 61.84 & \underline{66.25} & \underline{51.96} & 91.38 & \textbf{78.61} & 7.27 & 33.33 & 4.08 & 33.06 \\
Hiera \cite{hiera} & 60.38 & 45.71 & \underline{88.89} & 50.31 & 48.70 & 32.56 & \underline{96.55} & 68.69 & 17.86 & 71.43 & 10.20 & 38.97 \\
X-CLIP \cite{xclip} & \textbf{83.33} & \textbf{83.33} & 83.33 & \textbf{90.76} & 62.67 & 51.09 & 81.03 & 70.26 & \underline{36.92} & \underline{75.00} & 24.49 & \underline{48.25} \\
VideoMAE v2 \cite{videomaev2} & 71.79 & 66.67 & 77.78 & \underline{74.56} & 55.43 & 40.48 & 87.93 & 68.31 & \textbf{64.86} & \textbf{96.00} & \underline{48.98} & \textbf{70.08} \\
\end{tabular}
\end{adjustbox}
\vspace{2pt}
\begin{adjustbox}{width=0.75\linewidth}
\begin{tabular}{lcccccccccccc}
\midrule
 & \multicolumn{4}{c}{\textit{BWC Weapon Out}} & \multicolumn{4}{c}{\textit{Any Officer Handcuffing}} & \multicolumn{4}{c}{\textit{Other Officer Medical Treatment}} \\
\cmidrule(lr){2-5} \cmidrule(lr){6-9} \cmidrule(lr){10-13}
Model & F1 & P & R & AP & F1 & P & R & AP & F1 & P & R & AP \\
\midrule
\textit{Random Baseline} & 13.10 & 7.01 & \textbf{100.00} & 7.01 & 6.50 & 3.36 & \textbf{100.00} & 3.36 & 12.07 & 6.42 & \textbf{100.00} & 6.42 \\
CLIP \cite{clip} & \textbf{50.43} & \textbf{43.28} & 60.42 & 53.62 & 8.33 & \textbf{100.00} & 4.35 & \underline{48.82} & 16.95 & 33.33 & 11.36 & 33.28 \\
DINOv2 \cite{dinov2} & 41.79 & 32.56 & 58.33 & \textbf{57.87} & 25.00 & 44.44 & 17.39 & 40.00 & \underline{32.79} & \underline{58.82} & 22.73 & \underline{39.56} \\
Hiera \cite{hiera} & 26.15 & 16.04 & \underline{70.83} & 34.57 & \textbf{38.89} & 53.85 & 30.43 & 48.09 & 32.52 & 25.32 & \underline{45.45} & 27.87 \\
X-CLIP \cite{xclip} & \underline{44.44} & \underline{33.33} & 66.67 & \underline{54.31} & 34.48 & \underline{83.33} & 21.74 & \textbf{51.13} & 29.09 & \textbf{72.73} & 18.18 & \textbf{52.82} \\
VideoMAE v2 \cite{videomaev2} & 32.37 & 22.40 & 58.33 & 47.94 & \underline{37.84} & 50.00 & \underline{30.43} & 47.91 & \textbf{35.44} & 40.00 & 31.82 & 38.84 \\
\end{tabular}
\end{adjustbox}
\vspace{2pt}
\begin{adjustbox}{width=0.75\linewidth}
\begin{tabular}{lcccccccccccc}
\midrule
 & \multicolumn{4}{c}{\textit{Other Officer Physical Interaction}} & \multicolumn{4}{c}{\textit{Civilian Injured}} & \multicolumn{4}{c}{\textit{Civilian On Ground}} \\
\cmidrule(lr){2-5} \cmidrule(lr){6-9} \cmidrule(lr){10-13}
Model & F1 & P & R & AP & F1 & P & R & AP & F1 & P & R & AP \\
\midrule
\textit{Random Baseline} & 25.26 & 14.45 & \textbf{100.00} & 14.45 & 7.04 & 3.65 & \textbf{100.00} & 3.65 & 21.85 & 12.26 & \textbf{100.00} & 12.26 \\
CLIP \cite{clip} & \textbf{58.56} & \textbf{64.63} & 53.54 & \textbf{63.43} & 53.52 & 41.30 & 76.00 & \underline{69.82} & \textbf{60.55} & \textbf{49.25} & 78.57 & \textbf{73.30} \\
DINOv2 \cite{dinov2} & 54.55 & 51.82 & 57.58 & 55.95 & \underline{53.52} & \underline{41.30} & 76.00 & 57.89 & \underline{55.69} & 41.52 & \underline{84.52} & \underline{73.03} \\
Hiera \cite{hiera} & 40.54 & 27.68 & \underline{75.76} & 38.89 & 25.93 & 15.33 & \underline{84.00} & 31.96 & 29.30 & 17.83 & 82.14 & 54.38 \\
X-CLIP \cite{xclip} & 57.14 & \underline{57.73} & 56.57 & \underline{57.80} & \textbf{63.16} & \textbf{56.25} & 72.00 & \textbf{71.96} & 54.77 & \underline{42.04} & 78.57 & 72.60 \\
VideoMAE v2 \cite{videomaev2} & \underline{58.04} & 52.00 & 65.66 & 57.37 & 47.62 & 33.90 & 80.00 & 67.95 & 44.60 & 31.96 & 73.81 & 63.58 \\
\bottomrule
\end{tabular}
\end{adjustbox}
\end{center}